# ATG-MoE: Autoregressive trajectory generation with mixture-of-experts for assembly skill learning

Weihang Huang[a,†], Chaoran Zhang[a,†], Xiaoxin Deng[b,†], Hao Zhou[a], Zhaobo Xu[a], Shubo Cui[a], Long Zeng[a,*]

a Tsinghua Shenzhen International Graduate School, Tsinghua University, Shenzhen 518055, China

b Faculty of Engineering, Imperial College London, London SW7 2AZ, United Kingdom

* Corresponding author. E-mail address: zenglong@sz.tsinghua.edu.cn

† These authors equally contributed to this work.

**Abstract:**

Flexible manufacturing requires robot systems that can adapt to constantly changing tasks, objects, and environments. However, traditional robot programming is labor-intensive and inflexible, while existing learning-based assembly methods often suffer from weak positional generalization, complex multi-stage designs, and limited multi-skill integration capability. To address these issues, this paper proposes ATG-MoE, an end-to-end autoregressive trajectory generation method with mixture of experts for assembly skill learning from demonstration. The proposed method establishes a closed-loop mapping from multi-modal inputs, including RGB-D observations, natural language instructions, and robot proprioception to manipulation trajectories. It integrates multi-modal feature fusion for scene and task understanding, autoregressive sequence modeling for temporally coherent trajectory generation, and a mixture-of-experts architecture for unified multi-skill learning. In contrast to conventional methods that separate visual perception and control or train different skills independently, ATG-MoE directly incorporates visual information into trajectory generation and supports efficient multi-skill integration within a single model. We train and evaluate the proposed method on eight representative assembly skills from a pressure-reducing valve assembly task. Experimental results show that ATG-MoE achieves strong overall performance in simulation, with an average grasp success rate of 96.3% and an average overall success rate of 91.8%, while also demonstrating strong generalization and effective multi-skill integration. Real-world experiments further verify its practicality for multi-skill industrial assembly. The project page can be found at https://hwh23.github.io/ATG-MoE.

## 1 Introduction

As the manufacturing industry gradually shifts toward flexible production model characterized by high product diversity and small batch production [1], traditional robot systems relying on manual teaching and programming are facing growing challenges in adaptability. Whenever production tasks change, engineers need to replan the operation process and adjust the motion trajectories, which is both time-consuming and labor-intensive. Moreover, fixed trajectories are limited in handling potential variations in the task. In recent years, vision-guided grasping methods [2, 3] partly alleviate the dependence on precise calibration of object poses, enabling robots to perform operations under less restrictive conditions. However, their capability drops markedly when confronted with new task patterns or object categories. Therefore, a central challenge is to equip robots with a generalizable skill learning paradigm that allows them to autonomously acquire a set of fundamental manipulation skills and then flexibly combine them to accomplish a wide range of tasks, including previously unseen ones. This ability is crucial for achieving task-level flexibility.

In recent years, a variety of approaches have been proposed for robot skill learning, which can be broadly categorized into model-based learning, reinforcement learning (RL), and learning from demonstration (LfD) [4]. Although model-based learning and RL in principle offer certain advantages in policy optimization and generalization [5, 6], they often face practical difficulties in assembly scenarios, including the difficulty of model construction, slow policy convergence, and limited deployability. In comparison, LfD has become the most practical learning paradigm for assembly tasks because of its safety, precision, and high data efficiency. Among these approaches, kinesthetic teaching dominates industrial applications due to its high reproduction accuracy and strong consistency [7]. However, most existing strategies remain confined to force-control methods designed for specific tasks and targets, and therefore lack the ability to generalize to positional variation. Some studies introduce visual information to improve skill adaptability, but in most cases vision is only used for target localization or initial alignment, rather than for generating manipulation trajectories [8-12]. Other studies explore the combination of kinesthetic teaching and passive observation to support integrated multi-skill learning, with high-level task planning and low-level skill learning conducted in separate stages [13-15]. But the clear separation between stages requires independent training and increases system complexity, which prevents efficient coordination across the whole framework. Therefore, there is a pressing need for a new LfD method that can generalize across positional changes, unify perception and control, and support multi-skill integration.

To address the above challenges, this paper proposes an end-to-end autoregressive trajectory generation method with mixture of experts, namely ATG-MoE, for assembly skill learning. ATG-MoE is designed for assembly scenarios and aims to enable robots to flexibly respond to diverse task requirements through LfD. The method establishes a closed-loop learning pipeline that maps multi-modal inputs, including RGB-D images and natural language instructions, to trajectory outputs. The core contribution of ATG-MoE lies in the coordinated design of three modules. First, multi-modal feature extraction and fusion module improves the robot's integrated understanding of 3D spatial information and skill goals by encoding and combining visual perception with instructions, which allows the robot to respond appropriately to complex environmental states and human commands. Second, autoregressive trajectory generation module introduces a sequence modeling mechanism based on conditional probability, where subsequent actions are predicted from historical trajectories together with multi-modal scene information, thereby ensuring temporal continuity and dynamic adaptability in the generated trajectory. Third, multi-skill learning module adopts a mixture-of-experts architecture, in which a skill-specific routing network learns expert combinations tailored to different skills, while shared experts support the learning of common knowledge. This design balances skill specialization and shared knowledge, and thus enables coordinated modeling and efficient invocation of multiple skills within a unified framework. Compared with conventional learning frameworks that separate stages and skills, ATG-MoE provides a new end-to-end method in which visual information directly contributes to trajectory generation, while also achieving integrated multi-skill learning. As a result, it substantially simplifies the system design process and improves both policy transfer efficiency and generalization ability.

We conduct comprehensive experiments on eight representative assembly skills selected from a pressure-reducing valve assembly task. These experiments are designed to systematically evaluate the proposed method in terms of execution accuracy, trajectory efficiency and safety, positional generalization, cross-skill generalization, multi-skill integration, and real-world deployment. The results show that ATG-MoE achieves the most balanced performance in simulation, with an average grasp success rate of 96.3% and an average overall success rate of 91.8% across 512 test cases. In real-world experiments, after sim-to-real adaptation the proposed method maintains strong performance across eight assembly skills. These results demonstrate that ATG-MoE is effective and practical for multi-skill industrial assembly.

The rest of the content is organized as follows. Section 2 reviews related work on robot learning for assembly and LfD. Section 3 elaborates on the implementation methods of each module of ATG-MoE, including multi-modal feature extraction and fusion, coarse-to-fine autoregressive trajectory generation, and multi-skill learning. Section 4 uses the assembly task of the pressure-reducing valve as a case study to showcase the versatility and usability of our method. Finally, Section 5 concludes the study with a summary.

## 2 Related work

### 2.1 Robot learning for assembly skill

Robot learning encompasses a range of learning paradigms in robotics [16]. It typically takes environmental perception as input and aims to enable robots to make autonomous decisions and generate actions for interacting with objects to accomplish complex tasks. Achieving this capability fundamentally depends on learning diverse skills so as to improve task-level flexibility [17]. This requirement is particularly important in modern flexible assembly systems, where robots need to adapt skill parameters to rapidly evolving product variants and thus maintain reliable operation in uncertain environments. Current robot learning methods can be broadly divided into three categories [4]: model-based learning, reinforcement learning (RL), and learning from demonstration (LfD).

**Model-based learning** explicitly leverages an environmental model, such as a kinematic or dynamic model, to accelerate learning or directly solve robot control problems [18]. When the environmental model is known or can be estimated, the state transition mechanism of the environment, such as transition probabilities or a deterministic approximation model, can be embedded into the learning framework. Starting from an initial state, the model can then predict all possible future states along a complete trajectory, which transforms decision-making into a trajectory optimization or optimal planning problem by optimizing the action sequence to maximize cumulative reward. For example, Tang [19] developed a three-point contact physical model based on force/torque (F/T) analysis and geometric model, enabling a robot to learn an auto-alignment skill that accelerates the assembly process. Specifically, this method derives closed-form relationships among the F/T signal, tilt angle, and positional offset through the three-point contact mechanics model. It then constructs an objective function from pre-calibrated experimental data and solves for the friction coefficient through non-convex optimization to minimize model estimation error. Based on the estimated tilt angle and positional offset, the method finally generates a two-step compensatory trajectory, thereby achieving high-precision peg-hole insertion under hybrid force control. Beyond general model-based prediction of future states, some methods introduce receding-horizon optimization for online control, which forms the core of **model predictive control (MPC)** [20, 21]. After using an explicit process model to predict system behavior and optimize the control sequence, MPC applies only the first control input to the real system. At the next time step, it repeats prediction and optimization based on the latest system measurements and updates the control sequence accordingly. For instance, Piccinelli et al. [22] combined linear MPC with a virtual energy tank to enable remote precision assembly using two seven-degree-of-freedom robotic arms. Through receding-horizon optimization, their method regulates force and trajectory in real time and improves the balance between stability and transparency. Zhang et al. [23] are the first to combine a linear Koopman dynamic model with MPC to track and regulate the tension generated by active end-effector twisting together with gripper pose, thereby enabling cable harness assembly with a single robot arm in confined spaces. Under real-time constraints, this method addresses the strongly nonlinear control problem associated with deformable objects. Although MPC performs well when an accurate model is available, its reliance on model accuracy limits its applicability in unknown environments, such as new assembly tasks.

In most industrial scenarios, it is difficult to establish an accurate system dynamics model, which makes model-free learning a more practical choice. Among these methods, **reinforcement learning** directly optimizes interaction behavior through cumulative reward without explicitly relying on a state transition model, thereby providing a general framework for robot control [24]. RL solves decision-making through the joint optimization of a policy function and a value function. The policy function $\pi(a|s)$ directly maps states to actions, whereas the value function $V(s)$ or $Q(s,a)$ evaluates long-term return. Together, they form the basis for solving a Markov decision process [25]. Three main learning objectives arise in this setting. The first is **policy gradient learning**, which directly optimizes the policy function through gradient ascent during interaction. Although this class of methods provides an unbiased optimization path in theory, its high variance caused by reliance on Monte Carlo sampling creates serious challenges in practice. In robot assembly tasks, where rewards are often sparse and the state space is complex, pure policy gradient methods usually require excessive interaction data to converge, which severely limits their engineering applicability. The second is **value function learning**, which avoids explicit policy optimization and instead solves for the optimal value function at the end of training, from which the policy is implicitly derived through the Bellman optimality equation. Wu et al. [26] proposed a RL method for robotic insertion skill learning based on a dueling deep Q network (DQN) framework. Their method maps 6D F/T signals during the assembly process to long-term value scores of discrete actions and then derives the policy implicitly through an argmax operation. To accelerate training, the method introduces demonstration-based pretraining and a dynamic adjustment mechanism for insertion step length. These designs effectively address sparse rewards and safety constraints in high-precision assembly. The third is **actor-critic learning**, which combines the above two in a hybrid optimization framework. The actor optimizes the policy online, while the critic estimates value and provides low-variance gradients. This class of methods balances immediate policy improvement and long-term value estimation. Xu et al. [6] proposed an improved deep deterministic policy gradient (DDPG) algorithm that enabled robots to learn assembly policies in real-world scenarios. Their method combines the output of a conventional force controller with the actor network to generate continuous assembly actions while maintaining action safety, which significantly reduces the risk of random exploration. At the same time, it replaces manually designed reward functions with a fuzzy reward system and introduces a feedback mechanism for dynamically adjusting exploration noise. Together, these designs address the problems of sparse rewards and low exploration efficiency in assembly tasks.

Although RL provides a general learning framework for robotic assembly, its practical deployment in industrial scenarios often faces several challenges. First, the learning cost is difficult to control. RL relies on interaction with the environment to collect data, yet precision assembly tasks allow very little tolerance for physical trial and error. Random policy exploration can easily cause collisions and damage to components, and each failure may lead to equipment downtime and material loss. Second, there is a sim-to-real gap. To reduce the risk of real-world exploration, reinforcement learning is often pretrained in simulation. However, assembly tasks usually involve contact forces across multiple scales, which are difficult to model accurately with current physics engines. As a result, performance often drops sharply after transfer from simulation to real systems, which requires additional engineering effort for tuning. Third, rewards are sparse. In assembly scenarios, successful reward signals are naturally sparse and usually appear only when the final assembly is completed. RL therefore requires extensive exploration to associate actions with rewards. Considering these limitations of model-based learning and reinforcement learning in industrial deployment, another model-free learning paradigm, learning from demonstration, offers clear advantages in avoiding modeling bottlenecks and reducing risk on real systems. The central idea of learning from demonstration is to formulate assembly skill learning as a supervised learning problem that can be solved with a small amount of data. Its main principles are introduced in detail below.

## 2.2  Learning from demonstration

Learning from demonstration (LfD) is a robot learning paradigm based on supervised learning [4]. It directly acquires decision policies by analyzing expert demonstration trajectories, and is therefore also referred to as imitation learning. Its core mechanism is to formulate the state-action mapping as a supervised learning problem, where state observations from expert demonstrations serve as inputs and the corresponding actions serve as labels, allowing a policy network to be trained by minimizing action prediction error. This paradigm significantly offers high data efficiency, since strong policies can be learned from only a small number of demonstrations. At the same time, demonstration trajectories are naturally confined to the safe operating region, which improves both behavioral reliability and safety. This differs fundamentally from the reward-driven mechanism of reinforcement learning, because LfD does not rely on trial-and-error interaction with the environment. This property makes it particularly suitable for complex tasks in which the target behavior is difficult to script and the reward function is not easy to define. Existing demonstration methods can generally be divided into three categories: kinesthetic teaching, teleoperation, and passive observation [27], as shown in **Fig. 1**.

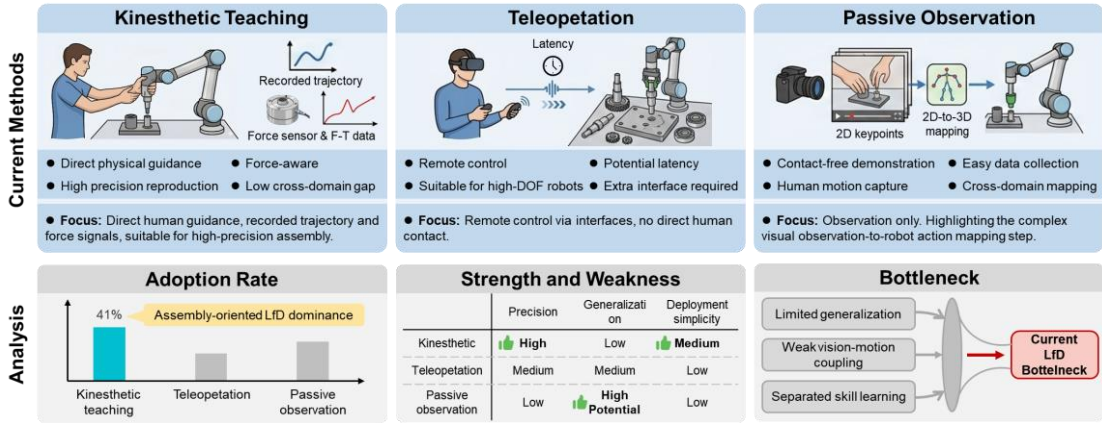

**Fig. 1.** Comparison of three types of LfD in robotic assembly.

**Kinesthetic teaching** is a LfD method in which motion trajectories are recorded by physically guiding the robot itself. Its main advantage lies in the intuitive demonstration process, together with the accurate reproduction of the demonstrated policy. For example, Wang et al. [28] proposed AL-ProMP for robotic polishing. By learning from demonstration trajectories and force signals in the time domain, their method directly associates force with spatial position, which enables force adaptation under varying speeds. The method also introduces a temporal scaling factor and a force scaling factor to improve skill generalization under changes in execution speed and tool replacement. Su et al. [29] proposed an assembly skill learning framework that combines kinesthetic teaching with adaptive impedance control. Their method synchronously collects trajectory and contact force data during demonstration, encodes skills with dynamical movement primitives (DMP), and generates reference trajectories and force profiles for new tasks through a combination of movement primitives (CoMP). It further incorporates an online adaptive impedance controller to achieve precise force tracking under unknown environmental stiffness, thereby enabling the assembly of non-cylindrical parts. **Teleoperation** is a LfD method based on a remote control interface, in which the operator commands the slave robot to perform tasks through force-feedback devices such as haptic controllers or master manipulators. Its main advantage is that it can support complex robots with high degrees of freedom, such as surgical robots, although it requires additional demonstration hardware and may suffer from control latency. Whitney et al. [30] developed a virtual-reality-based robot teleoperation system in which HTC Vive was used to control the viewpoint through head motion and to directly manipulate the robot end-effector through hand controllers. Their user study shows that the virtual reality interface improves the speed of cup-stacking demonstrations by 66% for novice users. **Passive observation** is a method that acquires expert behavior data through non-contact sensing such as vision or motion capture. Its greatest advantage is the ease of demonstration and its ability to capture high-degree-of-freedom motions of humans or complex mechanisms. However, since the observation domain differs from the robot execution domain, it requires complex cross-domain mapping to resolve the decoupling between observed motion and robot action. Deng et al. [31] proposed a passive observation framework that relies only on monocular video. Their method extracts 2D hand keypoints from video frames using Keypoint-RCNN and combines image processing with an iterative model based on a trust-region algorithm to map 2D observations to 3D manipulation trajectories without depth input, enabling a UR3 robot to successfully reproduce 15 object assembly tasks. After reviewing these three paradigms, it is worth noting that in the assembly domain, kinesthetic teaching has become the dominant LfD approach, accounting for 41% of reported methods [7]. It offers clear advantages in the precision and consistency of action reproduction, which makes it particularly suitable for assembly tasks with stringent trajectory accuracy requirements.

Although kinesthetic teaching has been widely adopted in assembly tasks, most existing studies focus on force-control tasks with relatively fixed start and end states. These tasks typically rely on F/T feedback control combined with demonstration trajectories to achieve stable and repeatable motion reproduction. However, because the task space is limited and environmental variation is small, the resulting policies often lack generalization ability and are therefore difficult to apply to the diverse operational demands of real industrial scenarios. To improve the adaptability of kinesthetic teaching, some studies have introduced visual information as an auxiliary cue [8-10]. For example, Ti et al. [11] used a vision system to obtain a coarse estimate of the socket pose for initial alignment. The visual localization error is compensated by compliant control to prepare for precise insertion under a three-point contact state. Nevertheless, F/T signals still serve as the primary basis for control, while the visual module only provides static information for initial positioning. Although this design extends the method's generalization ability to some extent, the weak coupling between vision and motion control prevents full exploitation of dynamic vision characteristics.

In addition, a small number of studies attempted to combine kinesthetic teaching with passive observation to support integrated multi-skill learning [13, 14]. For example, Ma et al. [15] used passive observation to learn task structure from human demonstrations, including target recognition and task sequencing, and performed high-level reasoning through a graph neural network. Kinesthetic teaching is used in the low-level skill execution module to learn multiple concrete skills such as picking and placing. By combining these two stages, their method forms a hierarchical task and motion planning framework with a clear separation between high-level reasoning and low-level execution. However, a clear boundary still remains between the two learning paradigms. Visual perception mainly serves the task recognition and decision branch, while the actual trajectory generation process does not fully exploit visual information. At the same time, training the two modules separately increases system complexity, which limits the efficiency and scalability of the method in real industrial deployment.

To overcome the above bottlenecks, we propose ATG-MoE, a LfD method for robot multi-skill learning. Built upon kinesthetic teaching, the proposed method further incorporates visual observations of the scene and natural language instructions to improve generalization across different scenarios and tasks, while also enhancing localization accuracy for assembly skills. On this basis, ATG-MoE adopts an end-to-end modeling that jointly handles visual perception and control within a single model. As a result, visual information directly contributes to trajectory generation, without requiring task recognition and trajectory generation to be treated as separate stages. Furthermore, ATG-MoE integrates the learning process of multiple assembly skills in a unified manner, thereby enabling integrated modeling and execution of multi-skill behaviors. The core design and implementation details of the proposed method are presented in the following sections.

# 3 Methodology

## 3.1 Overview

In industrial assembly scenarios, the completion of an assembly task usually requires multiple procedures and involves different types of skills. To enable a unified multi-skill learning, this paper constructs ATG-MoE, an integrated architecture that combines multi-modal understanding with autoregressive trajectory generation, as shown in **Fig. 2**. Taking multi-modal observations as input, the method achieves an end-to-end mapping from complex perception to precise trajectory through explicit geometric modeling and a skill-aware mixture-of-experts (MoE) mechanism. The overall design follows a coarse-to-fine spatial reasoning paradigm to generate trajectory gradually. Through multi-modal feature extraction and fusion, coarse-to-fine autoregressive trajectory generation, and MoE-based multi-skill learning, ATG-MoE provides a systematic solution for skill library construction in industrial environments.

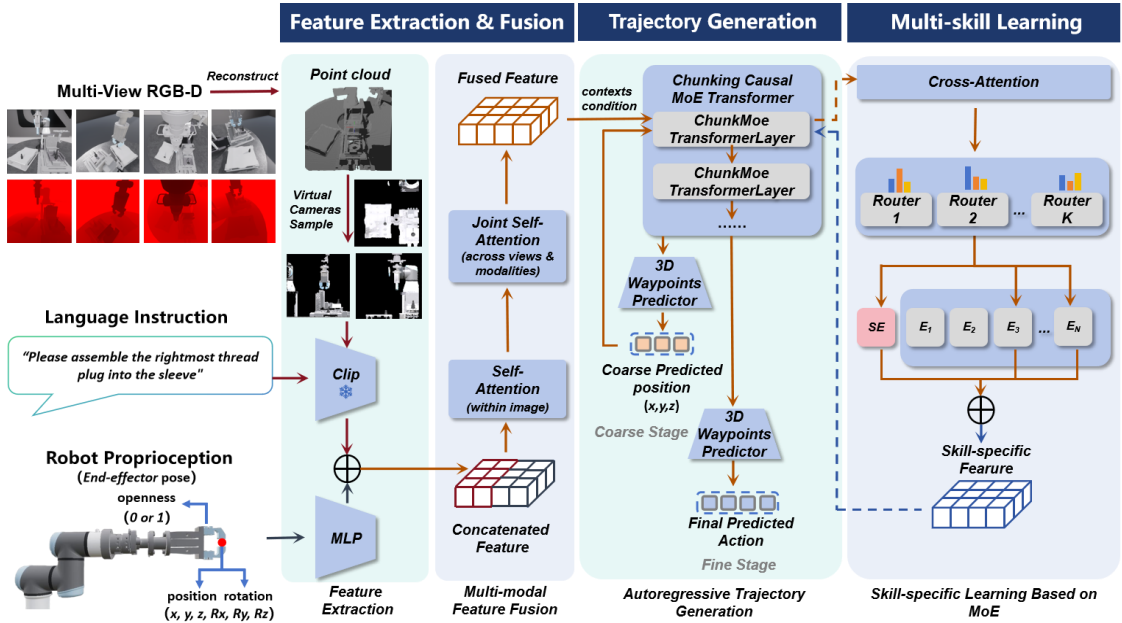

**Fig. 2.** Overview of ATG-MoE with its three main modules.

We formulate the problem as a multi-skill conditional LfD task. At the current time step $t$, the robot receives an observation $o_t$ from the environment, which includes multi-view RGB-D images, a natural language instruction, and robot proprioception. The objective of the model is to learn a conditional policy $\pi(a_t \mid o_t)$ that generates the end-effector action $a_t$ at conditioned on the observation $o_t$, thereby enabling the execution of assembly skills. In the feature extraction and fusion module, the system takes multi-modal inputs to jointly process visual perception and language-described skill information. These inputs serve two main purposes in ATG-MoE. First, the introduction of natural language instructions improves the robot's understanding of scene information and also allows users to interact with the robot more flexibly through language. Second, the method reconstructs point cloud from visual observations, which provides richer geometric cues for the assembly process. Through multi-view scene feature extraction, ATG-MoE tightly integrates the reconstructed 3D structure with

RGB visual cues. Compared with conventional methods that rely only on 2D image inputs, this design provides more accurate spatial information for target localization.

On this basis, ATG-MoE builds the trajectory generation network with an autoregressive model. The autoregressive trajectory generation module predicts the probability of each action conditioned on the historical trajectory and formulates a multi-step action sequence as a temporally dependent probability distribution through conditional probabilistic modeling. ATG-MoE adopts a unified encoder-decoder architecture to map the scene representation obtained from feature extraction and fusion into the action space of the robot end-effector. This design enables the generated trajectory to maintain strict temporal coherence while also supporting end-to-end generation.

ATG-MoE further enables collaborative multi-skill learning through a MoE module. We tightly integrate the MoE design with the autoregressive trajectory generation model, where each skill is associated with a dedicated router. According to the requirements of a specific skill, the router activates multiple experts and combines their outputs through skill-specific weights, thereby forming a specialized learning path for that assembly skill. Through this design, ATG-MoE is able to capture both shared knowledge across different skills and skill-specific expertise. As a result, it supports specialized learning and coordinated invocation of multiple assembly skills within a unified framework.

In the following, we describe the proposed ATG-MoE method in detail from three aspects: multi-modal feature extraction and fusion, coarse-to-fine autoregressive trajectory generation, and MoE-based multi-skill learning.

### 3.2 Multi-modal feature extraction and fusion

ATG-MoE adopts a hierarchical multi-modal feature fusion module that enables the joint encoding of heterogeneous data sources, including vision, language, and proprioception, through a multi-view Transformer architecture. As shown in **Fig. 2**, the method first extracts feature from each modality independently, then performs cross-modal interaction through spatial alignment and attention mechanisms, and finally produces a semantically guided joint representation.

#### 3.2.1 Multi-modal feature extraction

Given multi-view RGB-D images, natural language instructions, and robot proprioception, ATG-MoE uses independently designed encoders for different input modalities. Based on the depth input, the model back-projects image pixels into colored point clouds using the camera intrinsics. To unify the geometric semantics of multi-source information, the point clouds from all views are transformed into the robot base coordinate system through the camera extrinsics, concatenated into a unified point cloud representation, and cropped within a predefined workspace boundary. This process aligns RGB information with geometric structure in 3D space and ensures a consistent spatial reference across different views and skills. As a result, after reconstruction into the unified point cloud space, variations in camera position and even camera number across tasks have little effect on scene representation, which makes the model more robust to changes in viewpoint and sensing configuration.

Unlike directly processing sparse point clouds, ATG-MoE employs a differentiable point cloud renderer [32] to reproject the 3D point cloud into 2D images under a fixed set of viewpoints. They can be naturally embedded into a pretrained CLIP encoder [33] and transformed into a sequence of patch-level tokens $F_v^{patch}$ through block convolution. Specifically, each image is divided into $P \times P$ patches. They are projected into feature vectors with $C$ channels, yielding a patch-level representation $F_v^{patch} \in R^{(\frac{H}{P} \times \frac{W}{P} \times C)}$, where $H$ and $W$ represent height and width of the image. This operation maps the image from pixel space into a compact semantic space while preserving its spatial structure. The features from all views are then flattened and reorganized into a serialized visual representation $F_{vis} \in R^{(V \times N_P \times C)}$, where $N_p = \frac{H}{P} \times \frac{W}{P}$ denotes the number of patches in a single view.

For instruction processing, the input is first encoded by the CLIP text encoder into an embedding vector. A learnable projection matrix is then used to map the vector into the joint feature space shared with the visual representation, yielding a semantic query $F_{lang} \in R^{(L \times 2C)}$, where $L$ represents length of the embedding vector and $2C$ is the joint feature dimension. This dimension setting is introduced to match the dimensions of the subsequent joint representation of visual and proprioception.

For robot proprioception, state information such as the end-effector pose and gripper openness is projected by a multi-layer perceptron (MLP) into an embedding space, yielding the feature vector $F_{pro}$. The dimension of $F_{pro}$ is then broadcast to match that of the visual representation $F_{vis}$, so that it can participate in the subsequent self-attention computation. As a result, the model can establish position-wise associations between proprioception and visual context, thereby improving the physical feasibility of action generation.

After completing feature encoding for the three modalities, the model performs an initial fusion in the shared feature space. It first concatenates the visual features $F_{vis}$ with the proprioceptive feature vector $F_{pro}$ along the channel dimension, forming a joint perception-state representation $F_{v\_p} \in R^{(V \times N_P \times 2C)}$. A learnable positional encoding $P_{pos}$ is then added to this joint representation. The resulting feature sequence is subsequently concatenated with the language feature $F_{lang}$ along the length dimension, yielding the initial joint sequence $Z_{seq}$ that is fed into the transformer layers, as shown in **Eq. (1)**.

$$Z_{seq} = [F_{lang}; F_{v\_p} + P_{pos}] \in R^{(L+V \times N_p) \times 2C} \tag{1}$$

### 3.2.2 Cross-modal feature fusion

To fuse features from multiple modalities, ATG-MoE introduces a two-stage attention mechanism for fine-grained multi-modal integration. First, the model projects the joint sequence $Z_{seq}$ into the attention hidden dimension $d_{attn}$ through a linear layer, producing the projected sequence $Z_{proj} \in R^{(L+V \times N_p) \times d_{attn}}$. At this stage, to prevent spatial reasoning from being disturbed too early by global semantic information, $Z_{proj}$ is split along the sequence dimension into a language branch $F'_{lang} \in R^{(L \times d_{attn})}$ and a visual-proprioceptive branch $F'_{v\_p} \in R^{(V \times N_P \times d_{attn})}$. The language branch $F'_{lang}$ is temporarily bypassed, while $F'_{v\_p}$ undergoes local geometric association learning independently within each view. As shown in **Eq. (2)**, the model extracts the subsequence $F'^{patch}_v \in R^{(N_P \times d_{attn})}$ corresponding to view $v$, and then self-attention to obtain the locally enhanced feature $\widehat{F}'^{\text{patch}}_v$, where $W_Q, W_K, W_v$ denote learnable attention projection matrices, and $d_k$ is determined by the number of attention heads. This process enables the model to enhance responses to object boundaries and regions with similar texture patterns, while suppressing noise interference.

$$\widehat{F}'^{\text{patch}}_v = \text{Softmax}\left(\frac{\left(F'^{\text{patch}}_v W_Q\right)\left(F'^{\text{patch}}_v W_K\right)^\top}{\sqrt{d_k}}\right) F'^{\text{patch}}_v W_V + F'^{\text{patch}}_v \quad (2)$$

After local enhancement of visual feature, the model performs global cross-view and cross-modal interaction. Specifically, the locally enhanced joint features from all views, denoted as $\widehat{F}'_{v\_p}$, are concatenated again with the language feature $F'_{\text{lang}}$ to form the global feature $Z'_{\text{proj}}$. The model then performs joint self-attention in the unified feature space, allowing natural language instructions, visual information from different views and proprioception to adaptively allocate attention weights under a shared global receptive field, as shown in **Eq. (3)**.

$$\widehat{Z}'_{\text{proj}} = \text{Softmax}\left(\frac{(Z'_{\text{proj}} W_Q)(Z'_{\text{proj}} W_K)^\top}{\sqrt{d_k}}\right) Z'_{\text{proj}} W_V + Z'_{\text{proj}} \quad (3)$$

After global association is completed, the model explicitly removes the leading $L$ language tokens from the globally fused sequence $\widehat{Z}'_{\text{proj}}$ and retains only the visual-proprioceptive subsequence. This sequence is then passed through an output linear layer to restore the channel dimension to $2C$. Finally, it is rearranged into a tensor representation $Z_{\text{fusion}} \in \mathbb{R}^{V \times 2C \times \frac{H}{P} \times \frac{W}{P}}$, which restores the original view-wise spatial structure of the 3D scene.

### 3.3 Coarse-to-fine autoregressive trajectory generation

After obtaining the multi-modal fused feature $Z_{\text{fusion}}$, the model further infers the robot action at time step $t$. However, in a large 3D workspace, directly regressing high-precision continuous spatial coordinates in a single step leads to an excessively large search space and can easily overlook fine-grained local geometric contact cues. To address this issue, this paper proposes a two-stage coarse-to-fine trajectory generation, as shown in **Fig. 2**. It first performs coarse spatial localization from a global view to substantially narrow the search range, and then carries out position refinement and autoregressive expansion of heterogeneous actions on locally cropped features.

Before introducing the specific method, we first formulate assembly skill learning as a multi-dimensional conditional sequence generation problem. Unlike previous methods that predict the full action vector in parallel in a single step, this work generates action parameters autoregressively one chunk at a time, where each chunk contains multiple action dimensions, thereby explicitly modeling the physical and geometric conditional dependencies among different action dimensions. Specifically, at time step $t$, the complete action command is represented as an eight-dimensional vector, as shown in **Eq. (4)**, where $p_t$ denotes the three-dimensional position of the end effector. To avoid the periodic discontinuity introduced by a continuous Euler-angle representation, $r_t$ represents the rotational pose in quaternion form. The variable $g_t$ denotes the open-close state of the gripper.

$$a_t = (p_t, r_t, g_t) \tag{4}$$

To overcome the difficulty of directly inferring a high-dimensional continuous action vector, this work decouples heterogeneous actions in conditional distribution modeling. The translational component $p_t$, which requires extremely high spatial precision, is modeled as continuous-space maximum likelihood estimation based on multi-view heatmaps, and is further used as a spatial condition to guide the generation of subsequent rotational pose and gripper state. The remaining action parameters are discretized along the temporal axis and unfolded into an ordered token sequence of length $K$, denoted as $a_t^{\text{seq}} = (r_t^d, g_t)$, where $r_t^d$ represents the discretized rotation pose sequence. To balance autoregressive inference efficiency and the coupling among multi-dimensional action tokens, the sequence is divided into $M$ chunks, and each token in the sequence is assigned a chunk identifier $c_k \in \{1, \ldots, M\}$.

Based on the above definitions, at time step $t$, we decompose the joint conditional distribution at this step into a chunk-wise autoregressive form according to the probability chain rule, as shown in **Eq. (5)**. This formulation is supported by a clear physical rationale. In practical assembly operations, the end-effector pose often depends strictly on its precise spatial position $p_t$, while the gripper state further depends on the current approach pose. This autoregressive decomposition allows each subsequent decision to be inferred based on previously determined action chunks, and thus naturally captures the complex structured dependencies within the action.

$$P(a_t \mid Z_{\text{fusion}}) = P(p_t \mid Z_{\text{fusion}}) \prod_{m=1}^{M} P\left(a_{t,\{k|c_k = m\}}^{\text{seq}} \mid a_{t,\{i|c_i < m\}}^{\text{seq}}, p_t, Z_{\text{fusion}}\right) \tag{5}$$

### 3.3.1 Chunking causal MoE transformer

To estimate the conditional probabilities in **Eq. (5)**, this paper designs a Chunking Causal Mixture-of-Experts Transformer, abbreviated as CCMT, as a unified underlying module for sequential reasoning, as illustrated in **Fig. 3**. The forward pass of CCMT follows a strict logical chain of input embedding, sequence autoregression, and visual-conditioned querying. Its core computation process is described as follows.

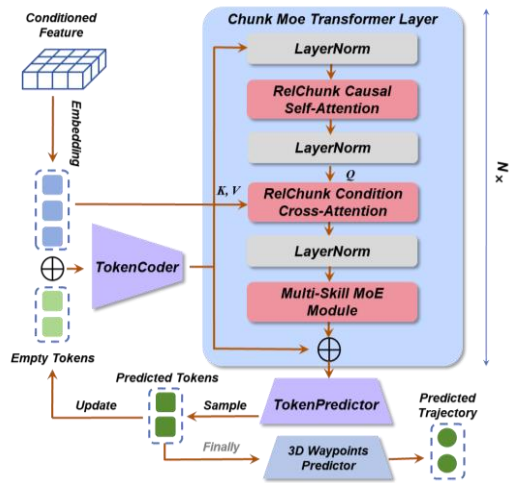

**Fig. 3.** Chunking Causal MoE Transformer (CCMT).

*Action type embedding and positional encoding.* At the input stage of action generation, the network maintains a learnable token type embedding matrix $E_{\text{type}}$ to distinguish the physical semantics of different dimensions in the action sequence, such as translation, rotation, and gripper state. This matrix is randomly initialized from the standard normal distribution $N(0,1)$ when the model is constructed. During autoregressive forward propagation, the model retrieves the semantic feature $E_{\text{type}}[id]$ from the matrix according to the predefined label id of each action dimension. In particular, for the continuous spatial localization dimension to be predicted, such as the spatial query token corresponding to $p_t$, the physical coordinates are not yet available. Its value embedding is therefore explicitly initialized as an all-zero placeholder, that is, $Embed_{\text{value}}(\cdot) = 0$. As a result, the initial input sequence feature $h_{\text{init}} \in R^{(L \times 2C)}$ fed into the transformer is mainly formed by the sum of the type semantics and the fixed absolute positional encoding $E_{\text{pos}}$, as shown in **Eq. (6)**. Here, $E_{\text{pos}}$ provides each token with global absolute position information in the sequence. For previously generated discrete historical actions, $val$ is mapped to the corresponding discrete value embedding.

$$h_{\text{init}} = Embed_{\text{value}}(val) + E_{\text{type}}[id] + E_{\text{pos}} \qquad (6)$$

*Causal self-attention with relative position bias.* To strictly preserve the autoregressive dependency among preceding action dimensions in **Eq. (5)**, the initial sequence $h_{\text{init}}$ is then fed into a self-attention layer with a lower-triangular causal mask. Specifically, $h_{\text{init}}$ is first projected through linear projection matrices into the query matrix $Q_{\text{self}}$, the key matrix $K_{\text{self}}$, and the value matrix $V_{\text{self}}$ required for self-attention. Given the strong local physical correlations between adjacent action dimensions, this work further introduces a relative distance bias matrix $B^{(\text{rel})}$ into the attention score computation. Let $q_i$ and $k_j$ denote the $i$-th row of $Q_{\text{self}}$ and the $j$-th row of $K_{\text{self}}$, respectively. Under the constraint of the causal mask, where $i \geq j$, each element of the self-attention score matrix is formulated as shown in **Eq. (7)**. Here, $B^{(\text{rel})}_{i-j}$ is a learnable relative position bias that depends only on the relative sequence distance between the current target token $i$ and the context token $j$. After the score matrix passes through the softmax operation, it is multiplied by the value matrix $V_{\text{self}}$ to produce the hidden state sequence $h_{\text{self}}$ after autoregressive interaction, as shown in **Eq. (8)**. This design gives the autoregressive process stronger translation invariance over the sequence and allows the model to focus on the relative conditional dependencies among action dimensions.

$$\text{Attn}^{\text{self}}_{i,j} = \begin{cases} \dfrac{q_i k_j^{\text{T}}}{\sqrt{d_k}} + B^{(\text{rel})}_{i-j}, & i \geq j \\ -\infty, & i < j \end{cases} \qquad (7)$$

$$h_{\text{self}} = \text{Softmax}\left(\text{Attn}^{\text{self}}_{i,j}\right) V_{\text{self}} \qquad (8)$$

*Cross-attention for visual intent querying.* After completing the autoregressive interaction within the action sequence, $h_{\text{self}}$ then attends to external visual features. Specifically, $h_{\text{self}}$ is projected as the query matrix $Q_{\text{cross}}$ of the cross-attention layer, while the external visual feature, such as $Z_{\text{fusion}}$, is flattened into a sequence $Z_{\text{f\_s}}$ and then projected into the key matrix $K_{\text{cross}}$ and the value matrix $V_{\text{cross}}$, as shown in **Eq. (9)**. Through this cross-modal interaction, tokens originally defined in the action space can fully absorb scene semantics and be decoded into intention query vectors $q$ enriched with visual guidance, as shown in **Eq. (10)**. In addition, the conventional feed-forward network at the end of the decoder is replaced with a MoE module, which improves multi-skill representation capability through dynamic routing. The detailed network structure and routing mechanism are described in the Section 3.4.

$$Q_{\text{cross}} = h_{\text{self}} W_Q^{\text{cross}} \quad K_{\text{cross}} = Z_{\text{f\_s}} W_K^{\text{cross}} \quad V_{\text{cross}} = Z_{\text{f\_s}} W_V^{\text{cross}} \tag{9}$$

$$q = \text{MoE}\left(\text{Softmax}\left(\frac{Q_{\text{cross}} K_{\text{cross}}^{\text{T}}}{\sqrt{d_k}}\right) V_{\text{cross}}\right) \tag{10}$$

### 3.3.2 Coarse stage: Spatial proposal based on global multi-view heatmap

Based on the problem definition, the first step in **Eq. (5)** is to estimate the spatial probability $P(p_t \mid Z_{\text{fusion}})$. The model first instantiates the first-stage network, CCMT-1, for global coarse localization. It splits $Z_{\text{fusion}}$ along the view dimension $V$ into feature maps $Z_v$ for individual views. For the coarse translational position, CCMT-1 follows the computation pipeline defined in **Eqs. (6)** to **(10)** and extracts a coarse intention query vector $q_{\text{coarse}}$. The model then broadcasts $q_{\text{coarse}}$ over the spatial dimensions and performs element-wise correlation with the feature map $Z_v$ to activate the corresponding response feature $M_v$, as shown in **Eq. (11)**.

$$M_v = q_{\text{coarse}} \otimes Z_v \in \mathrm{R}^{2C \times \frac{H}{P} \times \frac{W}{P}} \tag{11}$$

The response feature $M_v$ is then upsampled to the image resolution through multiple transposed convolution layers, denoted as ConvUpsample, and converted by softmax into a 2D probability heatmap $H_v^{\text{coarse}} \in \mathrm{R}^{H \times W}$ for each view. Based on the camera projection function $\Pi^{(v)}$, the modal searches for the location with the maximum joint likelihood across views and solves for the coarse 3D operation point $p_{\text{coarse}}$, as shown in **Eq. (12)**. At this stage, we also extract the operation point from the ground-truth demonstration trajectory and render it into the ground-truth heatmap $\widehat{H}_v^{\text{coarse\_gt}}$, using 2D Gaussian. The coarse localization heatmap loss $L_{\text{s1\_mse}}$ is computed using mean squared error (MSE), as shown in **Eq. (13)**.

$$p_{\text{coarse}} = \arg\max_p \sum_{v=1}^{V} \log H_v^{\text{coarse}} \Pi^{(v)}(p) \tag{12}$$

$$L_{\text{s1\_mse}} = \sum_{v=1}^{V} \text{MSE}\left(H_v^{\text{coarse}}, \widehat{H}_v^{\text{coarse\_gt}}\right) \tag{13}$$

### 3.3.3 Fine stage: Autoregressive generation of action sequences

After obtaining $p_{\text{coarse}}$, the model instantiates the second-stage network, CCMT-2, to perform local position refinement and autoregressive expansion of the action sequence. Specifically, the model centers a crop on $p_{\text{coarse}}$ and introduces a scaling factor $\alpha$ to crop and magnify the point cloud, which is then re-encoded into a high-resolution local feature representation, denoted as $Z'_{\text{fusion}}$. CCMT-2 reuses the heatmap prediction pipeline defined in **Eqs. (6)** to **(12)**, outputs a fine query vector $q_{\text{fine}}$ and predicts the fine heatmap $H_v^{\text{fine}}$, thereby estimating the precise 3D end-effector position $p_t$. In the same way, a fine localization loss is produced, as shown in **Eq. (14)**.

$$L_{\text{s2\_mse}} = \sum_{v=1}^{V} \text{MSE}\left(H_v^{\text{fine}}, \widehat{H}_v^{\text{fine\_gt}}\right) \tag{14}$$

After obtaining the spatial prerequisite $p_t$ in **Eq. (5)**, the model begins to autoregressively generate the discrete sequence $a_t^{seq}$. To construct fine-grained visual conditions for the autoregressive network, the model uses grid sampling to extract local features. Specifically, for the sub-pixel floating-point coordinates obtained by projecting $p_t$ onto the feature map, the grid sampling operation smoothly extracts a local spatial feature vector from $Z'_{fusion}$ by bilinearly interpolating the features of the four nearest neighboring integer pixels. At the same time, the model applies spatial pooling to the feature map to preserve global context. The concatenation of these two parts forms the conditional prompt sequence $h_{prompt}$, as shown in **Eq. (15)**. This $h_{prompt}$ replaces $h_{init}$ in CCMT and is directly fed into the causal self-attention layer.

$$h_{prompt} = [\text{GridSample}(Z'_{fusion}, p_t); \text{Pooling}(Z'_{fusion})] \qquad (15)$$

The visual prompt sequence $h_{prompt}$ is used as a conditional prefix and placed at the beginning of the action sequence to be generated. Under the lower-triangular causal mask and the relative distance bias, the network autoregressively predicts the discrete rotation pose and gripper state one chunk at a time based on $h_{prompt}$ and the previously generated action chunks. For the *m-th* chunk in the action sequence, the corresponding probability distribution is obtained by mapping the hidden state $h_{t,m}$ through the linear classification head $W_m^{cls}$, as shown in **Eq. (16)**.

$$P\left(a_{t,\{k|c_k=m\}}^{seq} \middle| a_{t,\{i|c_i<m\}}^{seq}, p_t, Z'_{fusion}\right) = \text{Softmax}\left(h_{t,m} W_m^{cls}\right) \qquad (16)$$

During training, to prevent error accumulation at the early stage of sequence generation, CCMT-2 adopts a teacher-forcing strategy and uses ground truth labels for parallel supervision of the entire action sequence. The model directly uses the ground-truth fine operation point $p_t^{gt}$ to perform grid sampling and construct $h_{prompt}$. It then maps the complete ground-truth discrete action sequence $a_t^{seq\_gt}$ from the demonstration trajectory into feature embeddings and concatenates them with $h_{prompt}$ before feeding them into CCMT-2, rather than generating actions step by step as in inference. Based on the predicted discrete sequence, a multi-dimensional cross-entropy loss is computed over the entire sequence as the optimization objective for autoregressive generation, denoted as $L_{ar\_ce}$ in **Eq. (17)**. The final overall training loss $L_{total}$ is defined as the weighted sum of the coarse localization heatmap loss, the fine localization heatmap loss, the autoregressive cross-entropy loss, and an auxiliary loss $L_{aux}$ used to balance the expert load in the MoE module, as shown in **Eq. (18)**, where each $\lambda$ is a hyperparameter that controls the balance among different training objectives. The auxiliary term $L_{aux}$ is further specified in Section 3.4.

$$L_{ar\_ce} = \sum_{k=1}^{K} \text{CE}\left(P\left(a_{t,\{k|c_k=m\}}^{seq}\right), a_{t,\{k|c_k=m\}}^{seq\_gt}\right) \qquad (17)$$

$$L_{total} = \lambda_1 L_{s1\_mse} + \lambda_2 L_{s2\_mse} + \lambda_3 L_{ar\_ce} + \lambda_4 L_{aux} \qquad (18)$$

### 3.4 Multi-skill learning based on MoE

To enable the model to continuously acquire new skills without forgetting previously learned ones, and thereby support multi-skill integration, this paper designs a multi-skill learning module based on a MoE network. This module is integrated into the CCMT introduced in Section 3.3.1. By incorporating an MoE routing mechanism into the feed-forward block, the model achieves efficient skill decoupling and stronger specialization capability, as illustrated in **Fig. 4**.

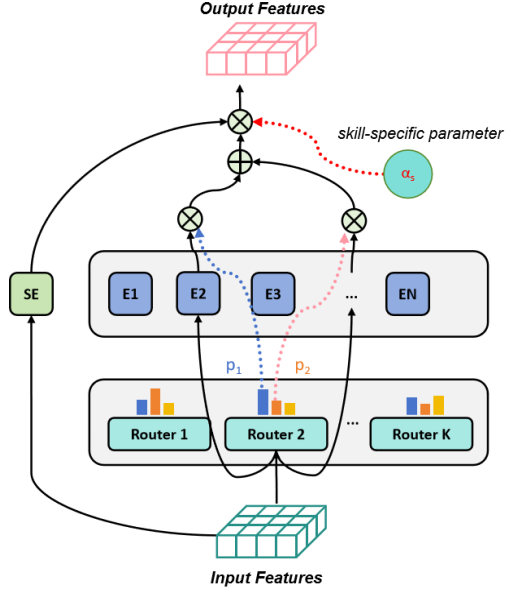

**Fig. 4.** Multi-skill learning module based on MoE.

To improve both knowledge sharing across skills and skill-specific specialization, ATG-MoE is built upon two key architectural designs. First, inspired by the strong performance of the DeepSeekMoE [34] in large language models, we design a group of shared experts (SE). These experts are implemented as MLPs, following the feed-forward design of the original transformer block. They are shared across all skills so that the model can learn common knowledge among different skills, thereby improving knowledge reuse and generalization. Second, in the design of the routing network, ATG-MoE departs from the conventional single-router architecture used in standard MoE and assigns an independent router to each operational skill. Conditioned on the skill identity and the input conditions, each router produces a weight distribution over experts for the current sample and dynamically determines which experts to activate, thereby providing a skill-specific routing path. This design allows the model to learn each skill in a more targeted manner. For expert selection, the router in ATG-MoE adopts the standard top-k strategy. For each input token, the router retains only the top $r$ experts with the highest probabilities, denoted as $E_e$, together with their corresponding weights $w_e$, while setting all remaining positions to zero, as shown in **Eq. (19)**. In this way, the model achieves sparse routing and keeps the computational cost under control.

$$y_{\text{router}}(x_{cond}) = \sum_{e=1}^{r} w_e E_e(x_{\text{cond}}) \quad (19)$$

Meanwhile, to prevent the MoE routing network from assigning most tokens to only a few experts during the early stage of training, which would cause only a small subset of experts to receive sufficient training, ATG-MoE introduces an auxiliary loss $L_{\text{aux}}$ to encourage a more balanced utilization of all experts. This loss is jointly determined by the average soft assignment probability $P_e$ and the actual hard assignment frequency $F_e$ of each expert $e$, as defined in **Eqs. (20)** to **(22)**. Specifically, let $N$ denote the total number of tokens over all samples and all dimensions, and let $n \in \{1, \ldots, N\}$ index these tokens. Here, $\pi_{n,e}^{\text{soft}}$ denotes the softmax probability that the *n-th* token is assigned to the *e-th* expert, while $\pi_{n,e}^{\text{hard}} \in \{0,1\}$ indicates whether the *n-th* token is actually routed to the *e-th* expert. In addition, $\tau$ is an empirical threshold chosen such that $L_{\text{aux}} = 0$ when $\sum_{e=1}^{r}(P_e F_e)$ reaches $\tau$.

$$P_e = \frac{1}{N}\sum_{n=1}^{N}\pi_{n,e}^{\text{soft}} \tag{20}$$

$$F_e = \frac{1}{N}\sum_{n=1}^{N}\pi_{n,e}^{\text{hard}} \tag{21}$$

$$L_{\text{aux}} = \frac{\sum_{e=1}^{r}(P_e F_e) - \tau}{r - 1} \tag{22}$$

Finally, to further strengthen differentiated learning across multiple skills, the model also learns the combination weight $\alpha_e$ between the output $y_{\text{router}}$ produced by the $r$ experts selected by the router and the output $y_{\text{SE}}$ of the shared experts. Under different skills, ATG-MoE learns different values of $\alpha_e$ during training, thereby mixing the specialized knowledge provided by the ordinary experts with the common knowledge encoded in the shared experts according to skill requirements, as shown in **Eq. (23)**.

$$\text{MoE}(x_{\text{cond}}) = \alpha_e y_{\text{router}}(x_{cond}) + (1 - \alpha_e)y_{\text{SE}}(x_{cond}) \tag{23}$$

## 4 Experiments

In this section, we evaluate ATG-MoE through a set of simulation and real-world assembly experiments using a real pressure-reducing valve as the case study. The experiments are designed to assess the feasibility, accuracy, and generalization capability of the proposed method under both simulated and practical deployment conditions. Video of all experiments is attached as supplementary material. Specifically, we aim to answer the following research questions:

**RQ1.** Can ATG-MoE achieve high accuracy, generating efficient trajectories with few execution steps, and improving operational safety?

**RQ2.** Can ATG-MoE exhibit strong positional generalization when the target object appears in varying positions?

**RQ3.** Can ATG-MoE generalize to unseen skills, thereby demonstrating cross-skill generalization capability beyond the skills observed during training?

**RQ4.** For industrial assembly tasks characterized by long-horizon operations, can ATG-MoE support integrated learning and coordinated invocation of multiple skills while maintaining strong performance?

**RQ5.** Can the proposed method be conveniently deployed on a real robotic platform, thereby enabling effective sim-to-real transfer for industrial assembly tasks?

### 4.1 Setup

This section introduces the pressure-reducing valve assembly task and experimental setup adopted in our study. The pressure-reducing valve assembly comprises 11 sequential processes [35], as shown in **Fig. 5**. Considering the representativeness of different operation types, and the feasibility of experimental implementation, we selected eight of these processes as representative skills for imitation learning and evaluation in our experiments. In the order of the original assembly workflow, these skills are:
- *[Sleeve Placement]*: This is the first half of process 1, which means placing the sleeve onto the fixture, waiting for the rubber ring assembly.
- *[Large Spring Insertion]*: It's process 3, which means inserting the large spring into the hole inside the sleeve.
- *[Rod Placement]*: This is also the first half of process 4, which means placing the rod onto the fixture, waiting for the rubber ring assembly.
- *[Rod Seating]*: Corresponding to process 5, this skill involves aligning the rod and further pressing it into the sleeve.
- *[Nut Seating]*: This is process 6, in which the nut is first aligned with the rod, and then seated onto the sleeve.
- *[Spring Insertion]*: In process 8, this operation refers to inserting the spring onto the pin-shaped cylinder at the top of the sleeve.
- *[Plug Seating]*: It's process 10, which means seating the plug onto the sleeve.
- *[Body Seating]*: It's process 11, which means seating the body onto the sleeve.

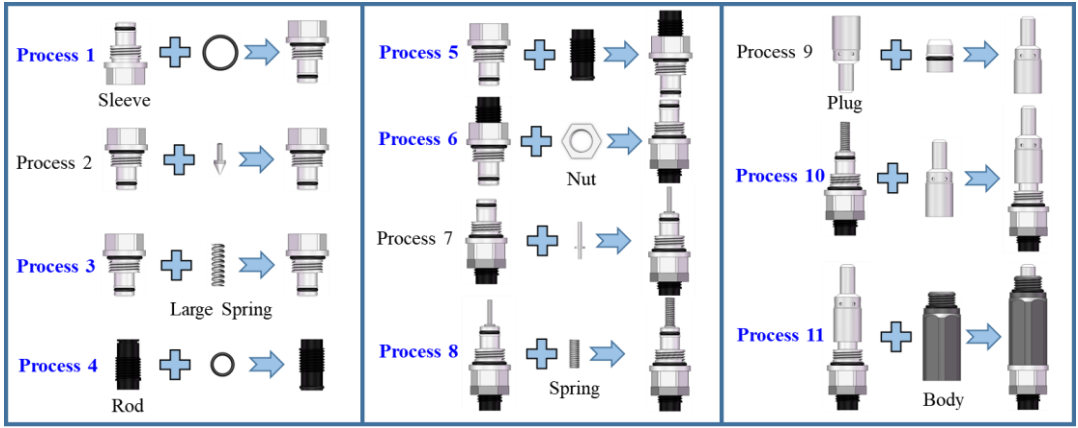

**Fig. 5.** Assembly process of pressure-reducing valve. Among the entire workflow, the eight processes highlighted in **blue** are selected as representative assembly skills for model training and evaluation. The names of all parts involved are also annotated in the figure.

As can also be observed from the naming of the above eight skills, we divide them into three categories, namely *Placement*, *Seating*, and *Insertion*, according to the level of execution precision required, with precision demands increasing in this order. Specifically, *Placement* skills mainly involve coarse placement of parts into the target region and therefore require relatively low precision. *Seating* skills require more accurate local alignment and pressing operations to bring a part into its final assembled state. *Insertion* skills usually involve constrained insertion into narrow mating structures, and thus impose the highest demands on positional accuracy and orientation control. This categorization provides a clear basis for evaluating the learning and generalization ability of the proposed method across skills of different difficulty levels.

As shown in **Fig. 6**, we establish a sim-to-real experimental platform for assembly skill learning. In the real-world setup, a UR3 robot equipped with a gripper is used to execute the learned assembly skills. For all skills, the manipulation starts from the part tray, where the target part is initially placed, and the subsequent assembly operation is performed on the fixture, which serves as the main workspace for part placement, seating, and insertion. To support perception during skill execution, a multi-view camera array composed of three Intel RealSense D435i cameras is deployed to capture RGB-D observations of the scene in real time, providing complementary visual information from different viewpoints. Correspondingly, a one-to-one simulation environment is constructed in Unity to faithfully reproduce the real assembly workspace, including the robot, part tray, and fixture. The robot motion in simulation is driven through URSim as a virtual robot controller, so that the ATG-MoE can send commands directly to the virtual controller to actuate the digital robot model, thereby enabling consistent training and evaluation across simulated and real-world settings.

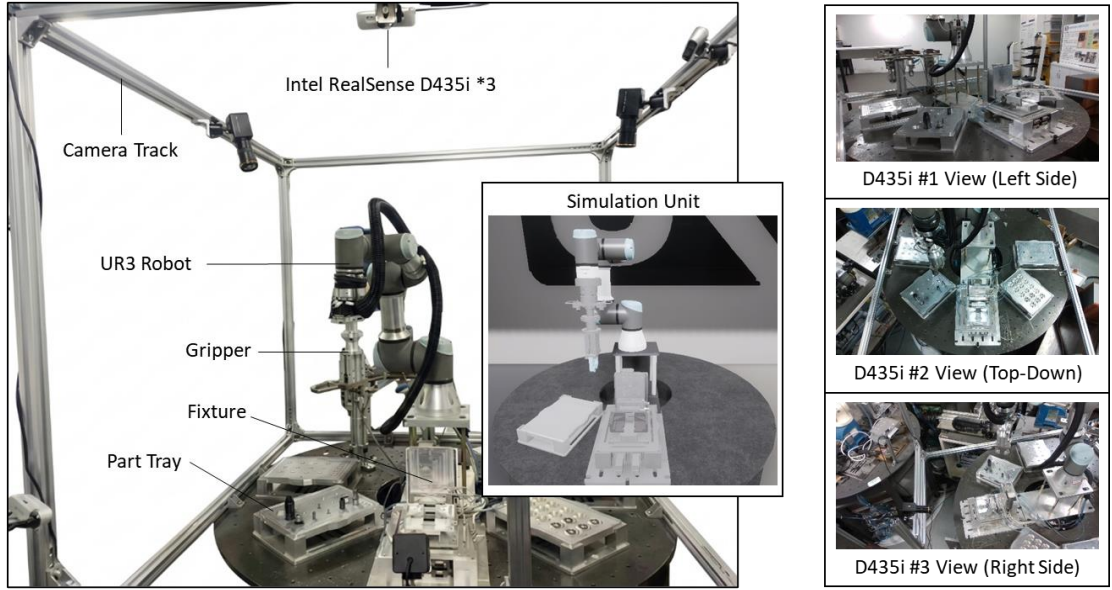

**Fig. 6.** A sim-to-real experimental unit for assembly skill learning. The real-world experimental setup with marked hardware components (**left**), the simulation setup constructed in Unity (**middle**), and RGB observations from three different viewpoints used as visual inputs to the model (**right**).

For both training and evaluation, each individual skill is conducted on a single NVIDIA GeForce RTX 4090 GPU, which is sufficient for the computational requirements of ATG-MoE. Unless otherwise stated, the model is trained for eight epochs with a batch size of 36 using the LAMB optimizer, and the initial learning rate is set to $5\times10^{-5}$. In addition, 2000 warm-up steps are used at the beginning of training to stabilize optimization and improve convergence. The fine stage in Section 3.3.3 uses a zoom scaling factor of 4, and data augmentation is applied during training to improve robustness and generalization. In the MoE module, the number of experts is set to 4, and top-1 routing is adopted, meaning that only the most relevant expert is activated for each input token, which helps balance model capacity and computational efficiency.

### 4.2 Training and evaluation

Based on the experimental setup described above, we carry out both simulation and real-world experiments for comprehensive evaluation. We first conduct sufficient experimental studies in simulation to verify the effectiveness and generalization ability of the proposed model, and then deploy it on the real robotic platform to further assess its practical applicability.

Through these results, we aim to answer the five research questions proposed at the beginning of Section 4.

### 4.2.1 Simulation experiments

In the Unity-based simulation environment, sufficient demonstration data can be conveniently collected for training. Each demonstration corresponds to the complete execution process of a single assembly skill by the robot. For ATG-MoE, however, trajectory generation does not require the entire demonstration sequence; instead, only five key frames are extracted from each demonstration, as shown in **Fig. 7**. These key frames can be regarded as milestones or sub-goals along the robot trajectory, and are defined as the robot states at which the end-effector velocity becomes zero, such as when the robot is about to perform grasping or assembly actions. For each key frame, RGB-D observations from four viewpoints are recorded together with the robot proprioceptive states. The five key frames, combined with the natural language instruction corresponding to the skill, form one complete demonstration sample. To enhance scene diversity and obtain positional generalization, the positions of the parts on the tray are randomized in simulation, such that each demonstration corresponds to a different object arrangement. Finally, 96 demonstrations are collected for each skill to train the model.

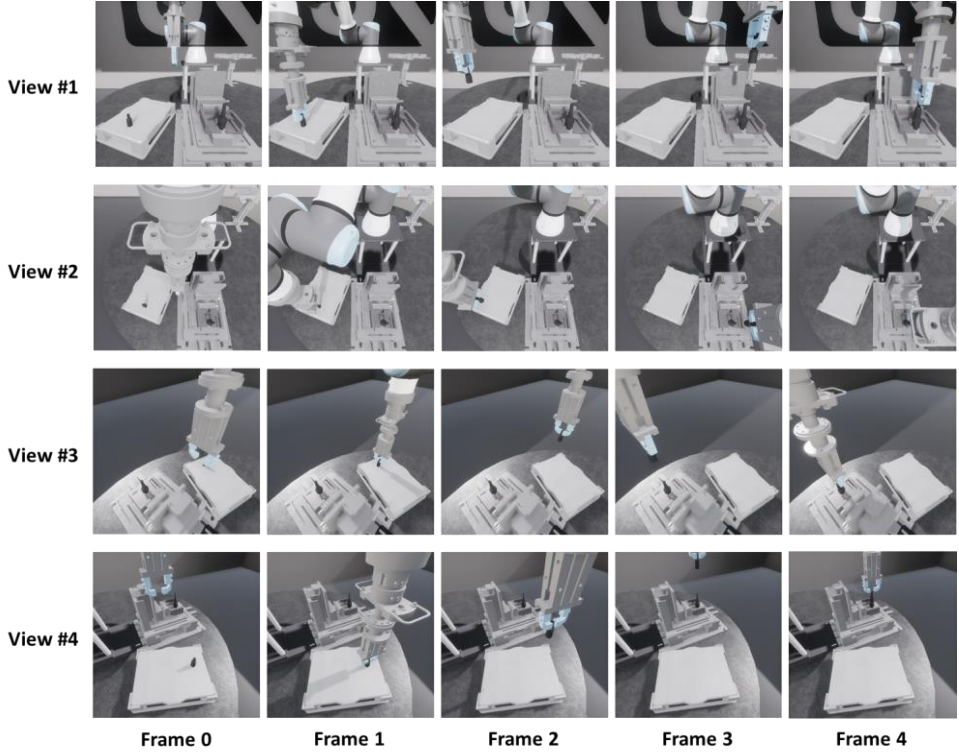

**Fig. 7.** Keyframe data for the *Plug Seating* skill in simulation. Five key frames are extracted from the demonstrated trajectory, and each frame contains observations from four camera views.

To evaluate the generalization capability of the trained model in a progressive manner, we design two levels of simulation experiments, namely Easy and Hard, as illustrated in **Fig. 8**. In the Easy setting, the tray position is kept the same as that used during training, while the part positions on the tray are randomly varied to generate 32 test cases. In the Hard setting, the degree of positional generalization is further increased by introducing four additional tray locations that are completely unseen during training. For each of these unseen tray locations, the part positions are further randomized eight times, resulting again in 32 test cases in total.

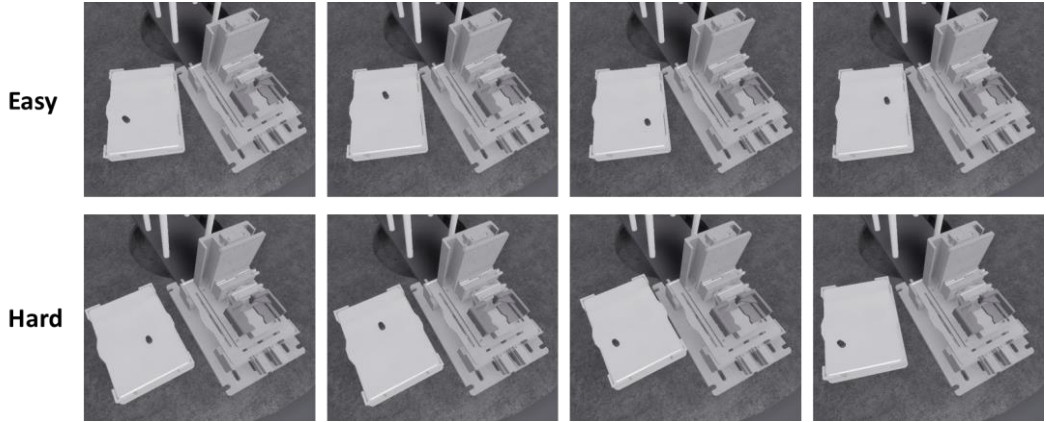

**Fig. 8.** Comparison of Easy and Hard evaluation settings using test cases of the *Rod Placement* skill. Each image corresponds to one case.

The comparative results in simulation are summarized in **Table 1**, where we compare ATG-MoE with several representative baselines, including Diffusion Policy (DP) [36], which models actions through iterative denoising, Action Chunking with Transformers (ACT) [37], which predicts action chunks in an autoregressive manner, and RVT-2 [32], a strong vision-based manipulation baseline that relies on transformer-based visual token modeling. To ensure a fair comparison, we slightly adapted the trajectory generation conditions of the baseline methods. Specifically, while the original settings typically use only the gripper open state or the current joint angles as input, we additionally provided the current-frame 6D end-effector pose together with the gripper openness, as the generation condition. This is equivalent to introducing one-step trajectory history and was necessary because, without this modification, the baseline methods performed very poorly in our assembly setting. Even under this more favorable and fair comparison protocol, the results show that ATG-MoE still achieves the most balanced and competitive performance in terms of success rate, safety, and efficiency, demonstrating its suitability for assembly skill learning.

**Table 1** Comparison of different methods in simulation experiments. Green-highlighted results indicate the best performance within each experimental category, while red-highlighted results indicate the worst. GSR (Grasp Success Rate) measures the success rate of grasping the target object in the first stage of a skill, whereas OSR (Overall Success Rate) measures the success rate of the complete skill execution, including both grasping and assembly.

| Methods | GSR (%) ↑ | | | OSR (%) ↑ | | |
|---|---|---|---|---|---|---|
| | Easy | Hard | Avg. | Easy | Hard | Avg. |
| DP [36] | 46.88 | 18.75 | 32.81 | 21.87 | 7.81 | 14.85 |
| ACT [37] | 62.89 | 62.11 | 57.42 | 35.16 | 25.78 | 30.47 |
| RVT-2 [32] | **96.88** | 85.27 | 91.07 | 72.32 | 61.61 | 66.97 |
| ATG-MoE | 96.10 | **96.49** | **96.30** | **92.19** | **91.41** | **91.80** |

|  | **Collision Rate (%)** ↓ | | | **Average Success Step** ↓ | | |
|---|---|---|---|---|---|---|
|  | Easy | Hard | Avg. | Easy | Hard | Avg. |
| DP [36] | **1.56** | 8.85 | 5.20 | 8.29 | 9.6 | 8.63 |
| ACT [37] | 18.75 | 18.75 | 18.75 | 6.32 | 7.97 | 7.07 |
| RVT-2 [32] | 1.79 | 16.07 | 8.93 | 7.98 | 6.73 | 7.55 |
| ATG-MoE | 4.30 | **3.91** | **4.10** | **5.02** | **5.18** | **5.10** |

To better understand the effectiveness of the proposed method, we further analyze the results from several perspectives. First, ATG-MoE achieves the highest overall success rate while maintaining a relatively low collision rate, which is particularly important for practical deployment in industrial environments. In our experiments, ATG-MoE reaches an average OSR of 91.8%, substantially outperforming the other baselines. This indicates that the proposed method is not only effective at grasping the target object, but also more reliable in completing the entire skill execution. At the same time, its collision rate remains low (4.1% on average), suggesting that the generated trajectories are safer and better aligned with the constraints of real assembly scenarios. The above observations are further supported by the trajectory visualization in **Fig. 9**. Taking the challenging *Large Spring Insertion* skill as an example, the trajectory generated by ATG-MoE is most consistent with the ground-truth, with both grasping and assembly key frames accurately aligned with the desired motion pattern. By contrast, although the other methods can also complete the skill in some cases, their generated trajectories exhibit more noticeable deviations from the best path, which may reduce execution safety in constrained industrial scenarios. The ability of ATG-MoE to generate such accurate trajectories mainly comes from its unified perception-and-control formulation: instead of separating target recognition and motion generation into loosely connected stages, the model directly generates trajectories conditioned on multi-modal observations, which reduces error accumulation across stages and leads to more reliable execution.

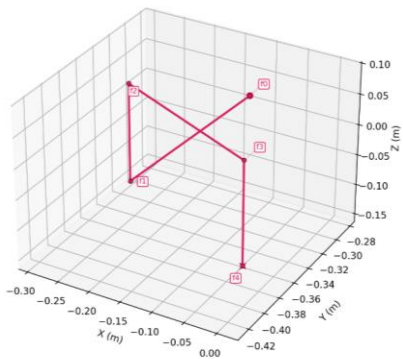

(a) Ground-truth

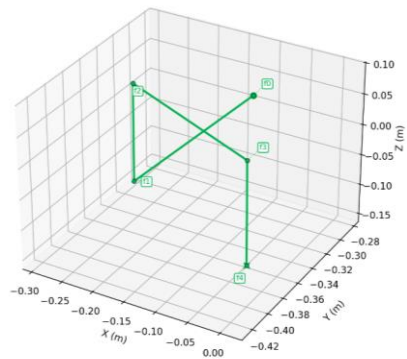

(b) ATG-MoE

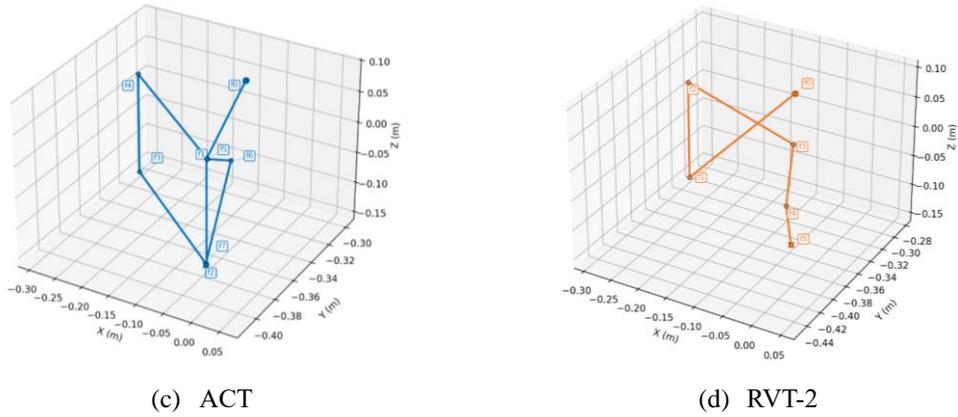

(c) ACT            (d) RVT-2

**Fig. 9.** 3D representation of successful robot end-effector's trajectory for the *Large Spring Insertion* skill. (a) shows the ground-truth trajectory, and (b)-(d) show trajectories generated by different methods. The marked points denote the key frames.

Second, ATG-MoE exhibits stronger positional generalization ability. Under the more challenging setting, where both the tray location and object arrangement vary beyond the training distribution, ATG-MoE still maintains a GSR of 96.49% and an OSR of 91.41%, showing only marginal performance degradation. This result suggests that the proposed method can make more effective use of visual information, rather than merely memorizing fixed workspace patterns. Because ATG-MoE operates on reconstructed scene geometry and fused visual features, it can adapt more naturally to changes in object position and scene layout.

Finally, ATG-MoE also demonstrates clear advantages in trajectory efficiency. It achieves the lowest average success step among all compared methods. This is also reflected in **Fig. 9**, where ATG-MoE generates the smallest number of key frames among all methods. Since the generated key frames correspond to the required execution steps, this result further indicates that ATG-MoE completes the skill with the highest trajectory efficiency. This efficiency can be attributed to the fact that ATG-MoE leverages keyframe-based demonstrations to learn the most informative intermediate sub-goals, enabling it to predict the next optimal trajectory point more directly and reach the target state with fewer redundant motions. In addition, our method does not require dense full-trajectory supervision, but only a sparse sequence of discrete key frames, which reduces the complexity of action generation.

**Table 2** provides a more detailed ATG-MoE's performance on each individual skill. Overall, the results show that ATG-MoE maintains consistently strong performance across all eight skills, indicating that its effectiveness is not limited to only a few simple operations but extends to a diverse set of industrial assembly behaviors. In particular, the seating skills show especially strong performance, indicating ATG-MoE's capability in precise local assembly. And *Rod Seating* is the most stable one, showing nearly perfect execution reliability, while *Large Spring Insertion* imposes the highest safety pressure, reflecting the greater contact sensitivity and collision risk of insertion-type operations. Building on this skill-level analysis, **Fig. 10** further compares different methods across all skills in a more detailed manner. The results show that the advantage of ATG-MoE is consistently maintained across a wide range of assembly skills. In particular, while some baseline methods already achieve competitive grasp success rates, their overall success rates drop markedly, indicating that they often fail in the subsequent assembly stage after successful object acquisition. By contrast, ATG-MoE maintains strong overall success across skills, showing that the proposed unified perception-and-control formulation supports grasping and the following placement, seating, and insertion operations more reliably.

Table 2 Comparison of ATG-MoE's performance across different skills in simulation experiments.

| Skill Type | Parts | GSR (%) ↑ | | | OSR (%) ↑ | | |
|---|---|---|---|---|---|---|---|
| | | Easy | Hard | Avg. | Easy | Hard | Avg. |
| Placement | Sleeve | 93.75 | 93.75 | 93.75 | 87.5 | 84.38 | 85.94 |
| | Rod | 93.75 | 100.0 | 96.88 | 87.5 | 100.0 | 93.75 |
| Seating | Rod | 100.0 | 100.0 | 100.0 | 100.0 | 100.0 | 100.0 |
| | Nut | 93.75 | 87.5 | 90.62 | 93.75 | 87.5 | 90.62 |
| | Plug | 100.0 | 100.0 | 100.0 | 93.75 | 81.25 | 87.5 |
| | Body | 96.88 | 96.88 | 96.88 | 93.75 | 93.75 | 93.75 |
| Insertion | Large Spring | 93.75 | 96.88 | 95.31 | 87.5 | 90.62 | 89.06 |
| | Spring | 96.88 | 96.88 | 96.88 | 93.75 | 93.75 | 93.75 |
| | | Collision Rate (%) ↓ | | | Average Success Step ↓ | | |
| | | Easy | Hard | Avg. | Easy | Hard | Avg. |
| Placement | Sleeve | 6.25 | 9.38 | 7.81 | 6.71 | 8.11 | 7.4 |
| | Rod | 12.5 | 0.0 | 6.25 | 5.11 | 5.03 | 5.07 |
| Seating | Rod | 0.0 | 0.0 | 0.0 | 5.22 | 5.28 | 5.25 |
| | Nut | 0.0 | 6.25 | 3.12 | 5.17 | 4.64 | 4.91 |
| | Plug | 0.0 | 0.0 | 0.0 | 4.0 | 4.19 | 4.09 |
| | Body | 0.0 | 0.0 | 0.0 | 4.43 | 4.87 | 4.65 |
| Insertion | Large Spring | 9.38 | 9.38 | 9.38 | 3.96 | 3.86 | 3.91 |
| | Spring | 6.25 | 6.25 | 6.25 | 5.57 | 5.47 | 5.52 |

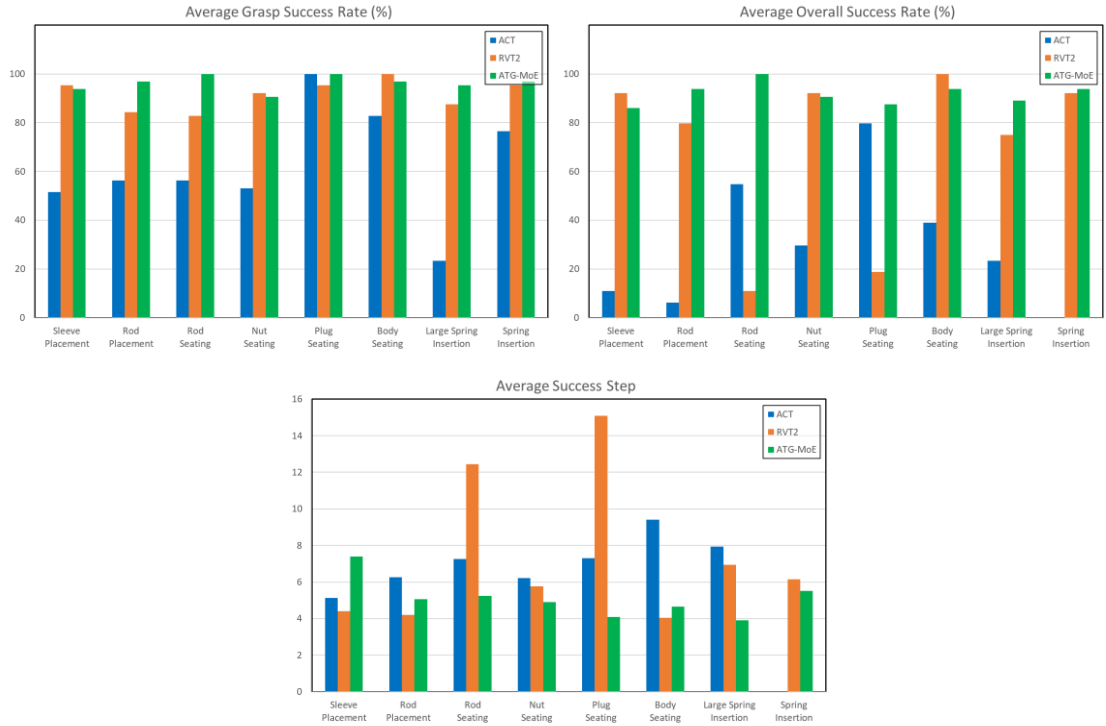

**Fig. 10.** Skill-wise comparison of ACT, RVT-2, and ATG-MoE in simulation experiments across three metrics: average grasp success rate, average overall success rate, and average success step.

The two heatmaps in **Fig. 11** are used to evaluate the cross-skill generalization capability of ATG-MoE. The results indicate that ATG-MoE does not merely memorize single-skill trajectories, but instead learns partially transferable perception–control patterns. Quantitatively, cross-skill generalization is stronger at the grasping stage than at the overall level: the average off-diagonal value is about 0.69 for grasp success rate, but drops to about 0.40 for overall success rate, showing that the main difficulty lies not in target perception and grasping, but in the subsequent fine assembly stage. In addition, skills of the same category are generally more transferable to each other. For example, in the grasp success rate heatmap, the average off-diagonal values within the same category reach about 0.74 for Placement, 0.73 for Seating, and 0.80 for Insertion, all higher than the overall off-diagonal average. However, cross-skill transfer is clearly asymmetric rather than mutual: many skill pairs exhibit strong one-way generalization, indicating that some source skills capture more generic control structures, while others mainly encode skill-specific strategies. Among all skills, *Rod Seating* is the most representative source skill, since it achieves the strongest outward transferability in the grasp-success heatmap, with an average off-diagonal value of about 0.95, suggesting that it learns the most universal skill representation. These findings also provide empirical support for subsequent multi-skill integration, as the clear knowledge-sharing potential, especially among similar skills, suggests that joint training could further improve the overall generalization capability of the model.

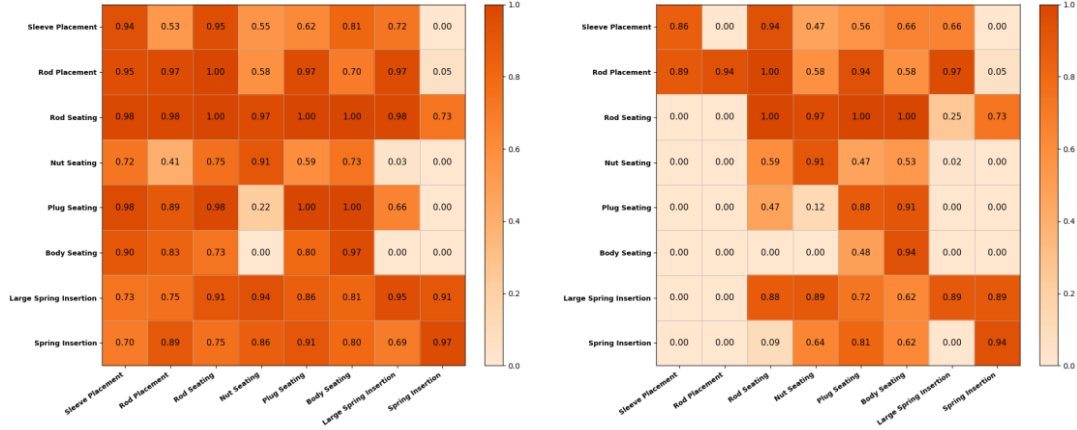

**Fig. 11.** Heatmaps for evaluating the cross-skill generalization capability of ATG-MoE. Each element $a[i,j]$ represents the performance of the model trained on skill $i$ and tested on skill $j$. The left heatmap shows the average grasp success rate, and the right heatmap shows the average overall success rate.

**Table 3** demonstrates the multi-skill learning capability of ATG-MoE. Under the S1+S2+S3 joint training setting, the model simultaneously achieves average values of 94.79% GSR and 90.62% OSR across the three skills. Compared with single-skill training, the jointly trained model also exhibits little evidence of catastrophic forgetting: the OSR values of S1, S2, and S3 remain close to their single-skill counterparts, and S1 even improves to 90.62% after joint training. These results directly show that a single model can maintain strong performance on multiple skills from different categories. This ability is closely related to the MoE design in our method. ATG-MoE can simultaneously capture shared knowledge across skills while preserving skill-specific specialization. In addition, the S1+S3 setting highlights the positive effect of knowledge sharing, yielding 100% average GSR and 92.96% average OSR on the two trained skills, while still achieving 98.15% GSR and 63.64% OSR on the unseen S2, which suggests that some skill combinations can promote more transferable representations. More importantly, from an application perspective, the multi-skill joint results provide support for long-horizon industrial assembly, where a unified model is expected to execute multiple skills sequentially rather than optimize a single isolated skill. This is further illustrated in **Fig. 12**.

**Table 3** Evaluation of the multi-skill learning capability of ATG-MoE using three skills from different categories. S1, S2, and S3 denote *Sleeve Placement*, *Body Seating*, and *Large Spring Insertion*, respectively. The model is trained on single-skill, two-skill, and three-skill combinations, and evaluated by grasp success rate and overall success rate. The blue-highlighted results correspond to the single-skill training setting and serve as the baseline for comparison with multi-skill training.

| *Sleeve Placement* | *Body Seating* | *Large Spring Insertion* | GSR (%) ↑ | | | OSR (%) ↑ | | |
|---|---|---|---|---|---|---|---|---|
| | | | S1 | S2 | S3 | S1 | S2 | S3 |
| √ | | | 93.75 | 81.25 | 71.88 | 85.94 | 65.62 | 65.62 |
| | √ | | 89.83 | 96.88 | 0.0 | 0.0 | 93.75 | 0.0 |
| | | √ | 81.25 | 73.44 | 95.31 | 62.5 | 0.0 | 89.06 |
| √ | √ | | 93.75 | 100.0 | 1.56 | 82.81 | 96.88 | 0.0 |
| √ | | √ | 100.0 | 98.15 | 100.0 | 95.31 | 63.64 | 90.62 |
| | √ | √ | 67.19 | 89.06 | 89.06 | 0.0 | 85.94 | 85.94 |
| √ | √ | √ | 95.31 | 96.88 | 92.18 | 90.62 | 93.75 | 87.5 |

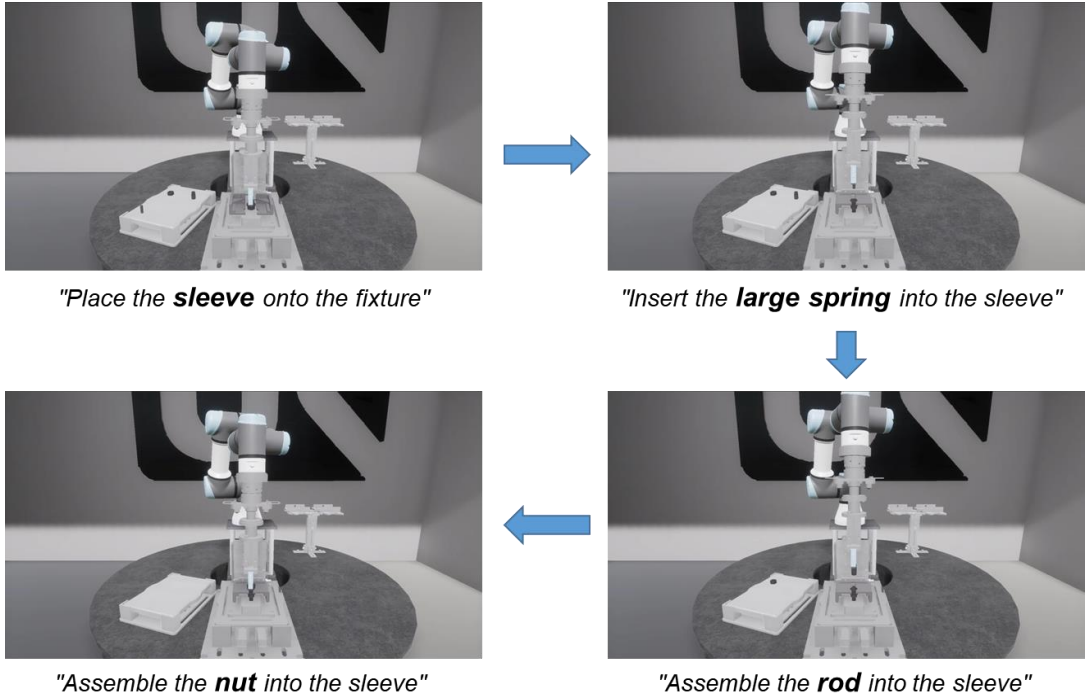

**Fig. 12.** Application scenario of multi-skill integration in the pressure-reducing valve assembly task. After multi-skill joint training, the model can continuously invoke four skills—*Sleeve Placement*, *Large Spring Insertion*, *Rod Seating*, and *Nut Seating*—under natural language instructions to accomplish part of the assembly process.

### 4.2.2 Real-world experiments

To transfer the policy from simulation to the real world, we adopt a sim-to-real pipeline initialized with the weights trained in simulation. For each skill, we manually collect 16 demonstration trajectories on the real robot, following a data format that is largely consistent with that used in simulation. These real-world demonstrations are then used to further fine-tune the model before deployment on the physical unit. During testing, we evaluate each skill on 10 different cases. In each case, the types and numbers of parts placed on the tray are varied to introduce visual distractions, thereby providing a more realistic evaluation of the model's robustness in practical assembly scenarios.

The execution process of three representative skills is illustrated through six key frames in **Fig. 13**. Frame 0 indicates the robot arm at its initial pose before execution. Frames 1–3 correspond to the grasping phase, including target alignment, object grasping, and lifting the part away from the tray. Frames 4–5 then show the assembly phase, where the robot moves the part to the target position and completes the corresponding placement, seating, or insertion operation. These examples qualitatively demonstrate that ATG-MoE can accomplish the full pipeline from grasp preparation to final assembly on the real robot.

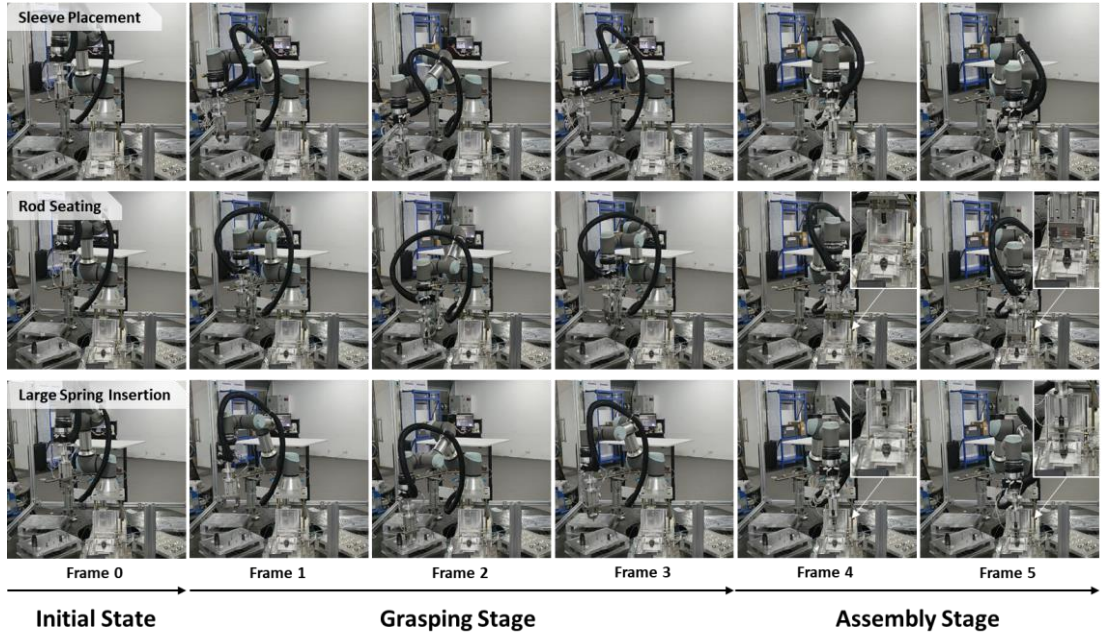

**Fig. 13.** Real-world execution sequences of three representative assembly skills.

The quantitative results in **Table 4** further validate the effectiveness of ATG-MoE across different skills in real-world experiments. Over eight evaluated skills, the model achieves consistently high grasp success rate, reaching 10/10 on most skills and remaining above 8/10 even for the more challenging insertion. The overall success rate also remains strong, with four skills achieving 10/10 and others reaching 8/10 or 9/10. Overall, these results show that ATG-MoE performs well across multiple assembly skills and can be robustly deployed on a physical robot unit under visually disturbed real-world conditions.

**Table 4** Comparison of ATG-MoE's performance across different skills in real-world experiments.

| Skill Type | Part | GSR ↑ | OSR ↑ |
|---|---|---|---|
| Placement | Sleeve | 9/10 | 8/10 |
|  | Rod | 10/10 | 8/10 |
| Seating | Rod | 10/10 | 9/10 |
|  | Nut | 10/10 | 10/10 |
|  | Plug | 10/10 | 10/10 |
|  | Body | 10/10 | 10/10 |
| Insertion | Large Spring | 8/10 | 8/10 |
|  | Spring | 10/10 | 10/10 |

## 4.3 Discussion and limitation

Overall, the results in Section 4 provide consistent answers to the five research questions. First, **Table 1** shows that ATG-MoE achieves the most balanced overall performance among the compared methods, confirming its effectiveness in industrial assembly skill learning. Second, the strong results under both Easy and Hard settings indicate that the model has good positional generalization and can remain robust under changes in object arrangement and scene configuration. Third, the cross-skill transfer results in **Fig. 11** suggest that ATG-MoE learns partially transferable representations, especially for grasping, although such transfer becomes weaker for the full assembly process. Fourth, the multi-skill results in **Table 3** and **Fig. 12** demonstrate that the proposed method can jointly learn multiple skills, while some skill combinations even improve transfer to unseen skills. Finally, the real-world results in **Fig. 13** and **Table 4** verify that the learned policy can be successfully transferred to the physical platform and maintain strong performance across different assembly skills under visually disturbed conditions. Taken together, these results validate ATG-MoE in terms of effectiveness, generalization, multi-skill integration, and real-world deployability.

Despite these encouraging results, the proposed method still has several limitations. The first limitation lies in the incomplete transferability across unseen skills, especially in the fine assembly stage. As shown in **Fig. 11**, although cross-skill transfer is relatively strong for grasping, the off-diagonal performance drops substantially at the overall level. This indicates that the model can generalize target perception and grasping more easily than the precise contact-rich motions. In other words, the current method still relies on skill-specific learning for high-precision assembly behaviors, and its zero-shot generalization to truly unseen skills remains limited. Future work could address this issue by incorporating F/T feedback and contact-aware control strategies to better handle fine contact variations during seating and insertion, or explicitly modeling reusable atomic skills that can be shared across different assembly operations.

The second limitation arises from the autoregressive trajectory generation itself. During training, ATG-MoE is optimized with teacher forcing, whereas during inference it predicts actions step by step based on its own previously generated outputs. This discrepancy introduces the well-known exposure bias problem, where small errors in earlier predictions may accumulate over time and gradually degrade the quality of the subsequent trajectory. Such error accumulation is particularly critical in assembly skills, because later-stage motions often require much higher precision than the initial aligning and grasping phases. Although the proposed model already improves temporal dependency modeling through action chunking, it still does not fundamentally eliminate the train–test mismatch. Future work could explore more robust sequence generation strategies, such as training schemes that explicitly account for rollout errors, or hybrid designs that combine autoregressive policy with online feedback correction during execution.

## 5  Conclusion

This paper presented ATG-MoE, an end-to-end method for assembly skill learning from demonstration that directly maps RGB-D observations and language instructions to manipulation trajectories through autoregressive trajectory generation and a mixture-of-experts design. By integrating perception, trajectory generation, and multi-skill learning into a unified model, the proposed method avoids the complexity of stage-wise pipelines and provides a practical solution for flexible robotic assembly. Experiments on eight representative assembly skills in a pressure-reducing valve assembly task showed that ATG-MoE achieves strong overall performance, with an average grasp success rate of 96.3% and an average overall success rate of 91.8%, while also demonstrating advantages in trajectory efficiency, safety, positional generalization, cross-skill transfer, multi-skill integration, and real-world deployment. We conclude by noting that, although the proposed method shows promising adaptability for multi-skill industrial assembly, it still faces limitations in fine contact-rich stages and autoregressive error accumulation, which motivates future work on more transferable skill representations and more robust feedback and correction mechanisms.

## CRediT authorship contribution statement

Conceptualization and methodology, Weihang Huang, Xiaoxin Deng; investigation, Chaoran Zhang; software, Shubo Cui; validation, Weihang Huang, Chaoran Zhang, Hao Zhou; writing-original draft, Weihang Huang, Chaoran Zhang; writing-review & editing, Xiaoxin Deng, Zhaobo Xu, Long Zeng; project administration, Long Zeng. All authors have read and agreed to the published version of the manuscript.

## Declaration of competing interest

The authors declare that they have no known competing financial interests or personal relationships that could have appeared to influence the work reported in this paper.

## Acknowledgement

This work was supported by National Natural Science Foundation of China (92467204) and Shenzhen Major Undertaking Plan (CKJZD20230923115503007).

## References

[1] Z. Xu, C. Zhang, S. Hu, Z. Han, P. Feng, L. Zeng, Reconfigurable flexible assembly model and implementation for cross-category products, Journal of Manufacturing Systems, 77 (2024) 154-169.
[2] L. Zeng, W.J. Lv, Z.K. Dong, Y.J. Liu, PPR-Net++: Accurate 6-D Pose Estimation in Stacked Scenarios, IEEE Transactions on Automation Science and Engineering, 19 (2022) 3139-3151.
[3] D.T. Huang, E.T. Lin, L. Chen, L.F. Liu, L. Zeng, SD-Net: Symmetric-Aware Keypoint Prediction and Domain Adaptation for 6D Pose Estimation In Bin-picking Scenarios, in: 2024 IEEE/RSJ International Conference on Intelligent Robots and Systems (IROS), 2024, pp. 2747-2754.
[4] Z. Liu, Q. Liu, W. Xu, L. Wang, Z. Zhou, Robot learning towards smart robotic manufacturing: A review, Robotics and Computer-Integrated Manufacturing, 77 (2022) 102360.
[5] C.C. Beltran-Hernandez, D. Petit, I.G. Ramirez-Alpizar, K. Harada, Learning to Grasp with Primitive Shaped Object Policies, in: 2019 IEEE/SICE International Symposium on System Integration (SII), 2019, pp. 468-473.
[6] J. Xu, Z. Hou, W. Wang, B. Xu, K. Zhang, K. Chen, Feedback Deep Deterministic Policy Gradient With Fuzzy Reward for Robotic Multiple Peg-in-Hole Assembly Tasks, IEEE Transactions on Industrial Informatics, 15 (2019) 1658-1667.
[7] V. Hernandez Moreno, S. Jansing, M. Polikarpov, M.G. Carmichael, J. Deuse, Obstacles and opportunities for learning from demonstration in practical industrial assembly: A systematic literature review, Robotics and Computer-Integrated Manufacturing, 86 (2024) 102658.
[8] F.J. Abu-Dakka, B. Nemec, J.A. Jørgensen, T.R. Savarimuthu, N. Krüger, A. Ude, Adaptation of manipulation skills in physical contact with the environment to reference force profiles, Autonomous Robots, 39 (2015) 199-217.

[9] Y. Ma, Y. Xie, W. Zhu, S. Liu, An Efficient Robot Precision Assembly Skill Learning Framework Based on Several Demonstrations, IEEE Transactions on Automation Science and Engineering, 20 (2023) 124-136.
[10] L. Rozo, A.G. Kupcsik, P. Schillinger, M. Guo, R. Krug, N. van Duijkeren, M. Spies, P. Kesper, S. Hoppe, H. Ziesche, M. Bürger, K.O. Arras, The e-Bike motor assembly: Towards advanced robotic manipulation for flexible manufacturing, Robotics and Computer-Integrated Manufacturing, 85 (2024) 102637.
[11] B. Ti, Y. Gao, M. Shi, J. Zhao, Generalization of orientation trajectories and force–torque profiles for learning human assembly skill, Robotics and Computer-Integrated Manufacturing, 76 (2022) 102325.
[12] M. Jiang, Q. Wang, H. Ai, Z. Dong, Y. Hu, A. Liang, Y. Wang, R. Li, Q. Liu, M. Chen, P. Búš, L. Zeng, Prompt2Act: Mapping prompts into sequence of robotic actions with large foundation models, Information Fusion, 127 (2026) 103923.
[13] S. Ji, S. Lee, S. Yoo, I. Suh, I. Kwon, F.C. Park, S. Lee, H. Kim, Learning-Based Automation of Robotic Assembly for Smart Manufacturing, Proceedings of the IEEE, 109 (2021) 423-440.
[14] Y. Gu, W. Sheng, C. Crick, Y. Ou, Automated assembly skill acquisition and implementation through human demonstration, Robotics and Autonomous Systems, 99 (2018) 1-16.
[15] R. Ma, J. Chen, J. Oyekan, A learning from demonstration framework for adaptive task and motion planning in varying package-to-order scenarios, Robotics and Computer-Integrated Manufacturing, 82 (2023) 102539.
[16] J. Peters, D.D. Lee, J. Kober, D. Nguyen-Tuong, J.A. Bagnell, S. Schaal, Robot Learning, in: B. Siciliano, O. Khatib (Eds.) Springer Handbook of Robotics, Springer International Publishing, Cham, 2016, pp. 357-398.
[17] C. Zhang, C. Zhang, Z. Xu, Q. Xie, P. Feng, L. Zeng, Embodied Intelligent Industrial Robotics: Concepts and Techniques, arXiv preprint arXiv:2505.09305, (2025).
[18] D. Nguyen-Tuong, J. Peters, Model learning for robot control: a survey, Cognitive Processing, 12 (2011) 319-340.
[19] T. Tang, Skill Learning for Industrial Robot Manipulators, in, 2018.
[20] C.B. Alba, Model predictive control, Springer Science & Business Media, 2012.
[21] Y. Tassa, T. Erez, E. Todorov, Synthesis and stabilization of complex behaviors through online trajectory optimization, in: 2012 IEEE/RSJ International Conference on Intelligent Robots and Systems, 2012, pp. 4906-4913.
[22] N. Piccinelli, R. Muradore, Linearized Virtual Energy Tank for Passivity-Based Bilateral Teleoperation Using Linear MPC, Ieee Transactions on Robotics, 41 (2025) 2589-2604.
[23] X. Zhang, H.C. Lin, Y. Zhao, M. Tomizuka, Ieee, Harnessing with Twisting: Single-Arm Deformable Linear Object Manipulation for Industrial Harnessing Task, in: 2024 International Conference on Intelligent Robots and Systems, Abu Dhabi, U ARAB EMIRATES, 2024, pp. 4069-4075.
[24] J. Kober, J.A. Bagnell, J. Peters, Reinforcement learning in robotics: A survey, International Journal of Robotics Research, 32 (2013) 1238-1274.
[25] L.P. Kaelbling, M.L. Littman, A.W. Moore, Reinforcement learning: A survey, Journal of artificial intelligence research, 4 (1996) 237-285.
[26] X.P. Wu, D.P. Zhang, F.B. Qin, D. Xu, Deep Reinforcement Learning of Robotic Precision Insertion Skill Accelerated by Demonstrations, in: 15th IEEE International Conference on Automation Science and Engineering (IEEE CASE), Univ British Columbia, Vancouver, CANADA, 2019, pp. 1651-1656.

[27] H. Ravichandar, A.S. Polydoros, S. Chernova, A. Billard, Recent Advances in Robot Learning from Demonstration, Annual Review of Control, Robotics, and Autonomous Systems, 3 (2020) 297-330.

[28] Y. Wang, C. Chen, F. Peng, Z. Zheng, Z. Gao, R. Yan, X. Tang, AL-ProMP: Force-relevant skills learning and generalization method for robotic polishing, Robotics and Computer-Integrated Manufacturing, 82 (2023) 102538.

[29] J. Su, Y. Meng, L. Wang, X. Yang, Learning to Assemble Noncylindrical Parts Using Trajectory Learning and Force Tracking, IEEE/ASME Transactions on Mechatronics, 27 (2022) 3132-3143.

[30] D. Whitney, E. Rosen, E. Phillips, G. Konidaris, S. Tellex, Comparing Robot Grasping Teleoperation Across Desktop and Virtual Reality with ROS Reality, in: N.M. Amato, G. Hager, S. Thomas, M. Torres-Torriti (Eds.) Robotics Research, Springer International Publishing, Cham, 2020, pp. 335-350.

[31] X. Deng, J. Liu, H. Gong, H. Gong, J. Huang, A Human–Robot Collaboration Method Using a Pose Estimation Network for Robot Learning of Assembly Manipulation Trajectories From Demonstration Videos, IEEE Transactions on Industrial Informatics, 19 (2023) 7160-7168.

[32] A. Goyal, V. Blukis, J. Xu, Y. Guo, Y.-W. Chao, D. Fox, RVT-2: Learning Precise Manipulation from Few Demonstrations, in: RSS 2024 Workshop: Data Generation for Robotics, 2024.

[33] A. Radford, J.W. Kim, C. Hallacy, A. Ramesh, G. Goh, S. Agarwal, G. Sastry, A. Askell, P. Mishkin, J. Clark, G. Krueger, I. Sutskever, Learning Transferable Visual Models From Natural Language Supervision, in: M. Marina, Z. Tong (Eds.) Proceedings of the 38th International Conference on Machine Learning, PMLR, Proceedings of Machine Learning Research, 2021, pp. 8748--8763.

[34] D. Dai, C. Deng, C. Zhao, R.x. Xu, H. Gao, D. Chen, J. Li, W. Zeng, X. Yu, Y. Wu, Z. Xie, Y.k. Li, P. Huang, F. Luo, C. Ruan, Z. Sui, W. Liang, DeepSeekMoE: Towards Ultimate Expert Specialization in Mixture-of-Experts Language Models, in, Association for Computational Linguistics, Bangkok, Thailand, 2024, pp. 1280-1297.

[35] Z. Xu, C. Zhang, S. Hou, Z. Han, L. Zeng, P. Feng, Assembly task planning framework based on knowledge graph, Journal of Intelligent Manufacturing, (2025).

[36] C. Chi, Z. Xu, S. Feng, E. Cousineau, Y. Du, B. Burchfiel, R. Tedrake, S. Song, Diffusion policy: Visuomotor policy learning via action diffusion, The International Journal of Robotics Research, 44 (2025) 1684-1704.

[37] T.Z. Zhao, V. Kumar, S. Levine, C. Finn, Learning fine-grained bimanual manipulation with low-cost hardware, arXiv preprint arXiv:2304.13705, (2023).